\RequirePackage{fix-cm}
\documentclass[twocolumn]{svjour3}          
\smartqed  
\usepackage{amsmath,amssymb} 
\usepackage{graphicx}
\usepackage{hyperref}
\usepackage{comment}

\usepackage{placeins}
\journalname{-}
\begin{document}
\title{Inferring bias and uncertainty in camera calibration}

\author{Annika Hagemann         \and
        Moritz Knorr \and
        Holger Janssen \and
        Christoph Stiller 
}

\institute{Annika Hagemann \at
	          Robert Bosch GmbH, Corporate Research, Computer Vision\\
	          Robert-Bosch-Strasse 200, 31139 Hildesheim, Germany\\ 
              \email{annika.hagemann@de.bosch.com}           
           \and
           Moritz Knorr \at
           Robert Bosch GmbH, Corporate Research, Computer Vision
            \email{moritzmichael.knorr@de.bosch.com}
           \and
          Holger Janssen \at
          Robert Bosch GmbH, Corporate Research, Computer Vision
           \email{holger.janssen@de.bosch.com}
          \and
           Christoph Stiller \at
           Karlsruher Institute of Technology (KIT)\\
           Institute of Measurement and Control\\
           \email{stiller@kit.edu}
}

\date{ }

\maketitle
\begin{abstract}
Accurate camera calibration is a precondition for many computer vision applications. 
Calibration errors, such as wrong model assumptions or imprecise parameter estimation, can deteriorate a system's overall performance, making the reliable detection and quantification of these errors critical.\\ In this work, we introduce an evaluation scheme to capture the fundamental
error sources in camera calibration: systematic errors (biases) and uncertainty (variance).  
The proposed bias detection method uncovers smallest systematic errors and thereby reveals imperfections of the calibration setup and provides the basis for camera model selection. 
A novel resampling-based uncertainty estimator enables uncertainty estimation under non-ideal conditions and thereby extends the classical covariance estimator. Furthermore, we derive a simple uncertainty metric that is independent of the camera model. 
In combination, the proposed methods can be used to assess the accuracy of individual calibrations, but also to benchmark new calibration algorithms, camera models, or calibration setups. We evaluate the proposed methods with simulations and real cameras.

\end{abstract}

\section{Introduction}
\label{sec:introduction}
\begin{figure*}[t]
	\centering
	\includegraphics[width=0.99\textwidth]{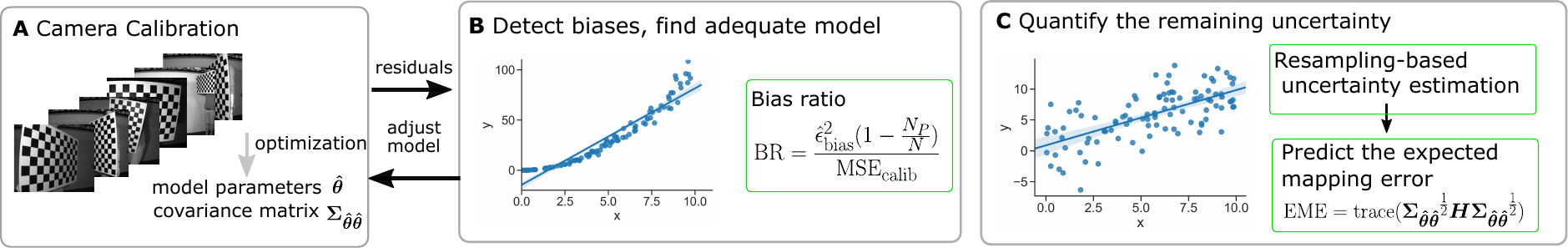}
	\caption{
		\textbf{Proposed evaluation scheme for camera calibration.} (A) Target-based camera calibration where intrinsic and extrinsic camera parameters are obtained through bundle adjustment. (B) Method to detect systematic errors (biases). The bias ratio (BR) quantifies the fraction of systematic error in the calibration residuals. (C) Quantification of the uncertainty. We propose a novel resampling-based method to estimate the covariance matrix of model parameters. Propagating the uncertainty to image space, the expected mapping error (EME) then provides a model-independent uncertainty metric. }
	\label{fig:overview}
\end{figure*}

Many applications in 3D computer vision rely on precisely knowing the camera's mapping from the 3D world to the 2D image.
This mapping $\boldsymbol{p}:\mathbb{R}^3\rightarrow\mathbb{R}^2$ is obtained during camera calibration.
Errors in the calibration will lead to false assumptions on the mapping, which can impact all subsequent inferences and deteriorate a system's overall performance~\cite{ozog_importance_2013,svoboda_what_1996,cheong_depth_2004,zucchelli_motion_2001,cheong_behaviour_2011,abraham_calibration_1998}. The detection and prevention of errors is therefore a critical aspect of the calibration.\par

To ensure high accuracy calibration, 
 recent advances have focused on easily applicable calibration solutions. 
 Starting with the seminal work by Zhang in 2000~\cite{zhang_flexible_2000}, 
 many contributions have been made with regard to improving and simplifying the calibration process~\cite{mei2007single,furgale2013unified,scaramuzza2006toolbox,matlab}. Multiple open-source toolboxes provide directly applicable functions for camera calibration, including the detection of calibration markers, different camera models and the complete bundle adjustment~\cite{ros,opencv}. Furthermore, several current approaches aim at tackling the problem of high uncertainty by developing systems for guided calibration~\cite{peng_calibration_2019,rojtberg_efficient_2018,richardson_aprilcal:_2013}. These systems infer most informative camera and calibration body constellations and guide users towards taking these images. Thus, both the data collection and the estimation process have been simplified substantially. \par 
Despite this progress in the calibration process, few have focused explicitly on tools to assess and interpret calibration results. One of the most common tools to assess the quality of a calibration remains to be the inspection of reprojection errors, as they are directly accessible after every calibration~\cite{matlab,opencv}. While residual errors reveal large systematic errors or unsuccessful optimizations, they cannot provide a full picture on the quality of the calibration. Such a full evaluation scheme would be critical to ensure high accuracy calibrations, but also to benchmark new calibration setups or solutions.\par 

In general, the error sources of camera calibration can be divided into systematic errors (bias) and uncertainty (variance)~\cite[p.116]{forstner2016photogrammetric}\cite{hagemann2020bias}. As in any data modeling problem, systematic errors occur if a chosen model is not sufficiently flexible or imposes false assumptions on the data (Fig.~\ref{fig:overview}B). In camera calibration, systematic errors can be caused by a camera projection model not being able to reflect the true geometric camera characteristics, e.g. if lens distortions are neglected. But also other sources, such as an uncompensated rolling shutter~\cite{oth2013rolling}, or non-planarity of the calibration target~\cite{lavest1998we} can result in systematic errors.\\
A high uncertainty, on the other hand, describes that model parameters could not be estimated reliably based on the available data. In general, uncertainties occur if the data is subject to noise and the amount of data is limited (Fig.~\ref{fig:overview}C). In camera calibration, high uncertainties are commonly caused by a lack of images used for calibration, bad coverage in the image, or a non-diversity in calibration target poses~\cite{sturm1999plane,peng_calibration_2019,rojtberg_efficient_2018,richardson_aprilcal:_2013}.\par

In this paper, we address the challenge of quantifying both, systematic errors and uncertainty, in target-based camera calibration. Our goal is to reliably detect and quantify both types of errors and to condense the result to easily interpretable measures.
Extending our recent work on evaluation metrics~\cite{hagemann2020bias}, we provide four
 main contributions (Fig.~\ref{fig:overview}): 

\begin{itemize}
	\item A method to \textbf{detect systematic errors} (biases) in camera calibration. The method is based on estimating the observational noise in marker detections and thereby disentangles random from systematic errors in the calibration residual.
	\item A method to \textbf{estimate parameter uncertainties under non-ideal conditions}. We show that the standard estimator for the covariance matrix underestimates the uncertainty under non-ideal conditions and propose a resampling approach for reliable estimation.
	\item A method to \textbf{predict the expected mapping error} (EME) in image space, which quantifies the uncertainty (variance) in model parameters in a model-independent way.
	\item A detailed \textbf{comparison of existing methods} to quantify bias and uncertainty. 
\end{itemize}
The main advantages of our approach towards quantifying systematic errors and uncertainty are that they are (a) applied as post-processing and thus build upon already captured data and (b) abstract from the underlying camera model, resulting in comparable results across different calibrations. 
We evaluate the proposed methods with both simulations and real cameras.

\section{Related Work}
\label{sec:stateoftheart}

\subsection{Detecting systematic errors}
Detecting systematic errors in camera calibration poses a challenge, because unlike the calibration of other measurement devices, there is typically no ground-truth device to compare with. Instead, camera characteristics are typically inferred indirectly through observations of well-known 3D objects. \par
The most common approach to detect systematic errors is by inspection of reprojection errors, i.e. the difference between predicted and observed image coordinates on the calibration dataset~\cite{matlab,opencv,beck_generalized_2018,schops_why_2019}. As the observational noise is typically assumed to be normally distributed, the reprojection errors (i.e. the residuals) should also follow a Gaussian distribution.
To detect systematic errors, the two-dimensional distribution of residuals can be visualized and deviations from the expected Gaussian distribution are an indicator for systematic errors~\cite{beck_generalized_2018}.\par To put this qualitative comparison into numbers, recent work proposed the KL-divergence between a 2D normal distribution, and the empirical distribution of reprojection error vectors as a measure of biasedness~\cite{schops_why_2019}. More precisely, it was proposed to compute the median KL-divergence over a 50x50 grid across the image, where low values indicate a large similarity to the Gaussian (low bias) and high values indicate dissimilarities (higher bias). \par
In addition to the distribution of residuals, the magnitude of residuals is a common indicator for systematic errors. A common approach is to compare the root mean squared error (RMSE) or reconstruction result against expected values obtained from earlier calibrations or textbooks~\cite{luhmann2013close}. However, the magnitude of residuals varies for different cameras, lenses, calibration targets, and marker detectors and therefore only allows capturing large errors in general.\par
Professional photogrammetry often makes use of precisely manufactured and highly accurate 3D calibration bodies~\cite{Rautenberg}. Images captured from predefined viewpoints are then used to perform a 3D reconstruction of the calibration body, where different length ratios and their deviation from the ground truth are compared against empirical data. While these methods are both, highly accurate and repeatable, they are often not feasible or too expensive for typical research and laboratory settings and require empirical data for the camera under test.

\subsection{Quantification of uncertainty}
The uncertainty of a calibration can, in general, be quantified by the (co-)variance in estimated model parameters $\boldsymbol{\hat{\xi}}$. As calibration is typically performed as a least squares estimation, the standard estimator for the covariance matrix $\hat{\boldsymbol{\Sigma}}_{\boldsymbol{\hat{\xi}}\boldsymbol{\hat{\xi}}, \mathrm{std}}$ is given by an approximated backpropagation of the observational noise in the data,
\begin{equation}
\hat{\boldsymbol{\Sigma}}_{\boldsymbol{\hat{\xi}}\boldsymbol{\hat{\xi}}, \mathrm{std}} = \hat{s}_d^2( \boldsymbol{J}_{\mathrm{calib}}^T  \boldsymbol{J}_{\mathrm{calib}})^{-1},
\end{equation}
where $\boldsymbol{J}_{\mathrm{calib}}$ is the Jacobian of calibration residuals and $\hat{s}_d^2 = \mathrm{MSE}_{\mathrm{calib}}/(1-\frac{N_P}{N})$ is the estimated accuracy, obtained from the mean squared error of the calibration $\mathrm{MSE}_{\mathrm{calib}}$, the number of parameters $N_P$ and the number of observations $N$~\cite[p.92-96]{luhmann2013close}~\cite[p.141-142]{hartley_multiple_2004}.\par
Since the covariance matrix is high dimensional and its interpretation is non-trivial, it is typically reduced to a scalar metric. Typical choices are the trace of the covariance matrix~\cite{peng_calibration_2019}, or the maximum index of dispersion~\cite{rojtberg_efficient_2018}. 
However, given the variety of camera models, from a simple pinhole model with only three parameters, up to local camera models with around $10^5$ parameters~\cite{beck_generalized_2018,schops_why_2019}, parameter variances are difficult to interpret and not comparable across camera models.\par
To address this issue, the parameter's influence on the mapping can be considered.
The metric \emph{maxERE}~\cite{richardson_aprilcal:_2013} propagates the parameter covariance to image space by means of a Monte Carlo simulation. More precisely, a 5x5 grid of 3D points is projected into the image with a sampled set of model parameters. This yields a distribution of image coordinates for each grid point. The value of \emph{maxERE} is then defined by the standard deviation of the most uncertain grid point. The \emph{observability} metric~\cite{strauss_kalibrierung_2015} uses an analytical approach and weights the uncertainty in estimated parameters with the parameters' influence on the mapping. This is achieved by linearly approximating the influence of parameter errors on a model cost function. Importantly, this model cost function takes into account that errors in intrinsic parameters can partially be compensated by a change of the camera coordinate system (i.e. by adjusting the extrinsics). 
The observability metric is then defined by an increase in calibration cost in the least observable parameter direction, where low observability values correspond to high uncertainty.\\
While both of these metrics provide valuable information about the uncertainty, there are some shortcomings in terms of \emph{how} uncertainty is quantified. The \emph{observability} metric does not consider the whole uncertainty, but only the most uncertain parameter direction. Furthermore, it quantifies uncertainty in terms of an increase in the calibration cost, which can be difficult to interpret. \emph{maxERE} quantifies uncertainty in image space and is thus easily interpretable. However, it relies on a Monte Carlo Simulation instead of an analytical approach and it does not incorporate potential compensations of errrors in the intrinsics by a change of the coordinate system.
\par
A final metric to assess the accuracy of a calibration are the reprojection errors on a test dataset~\cite{sun2006empirical,richardson_aprilcal:_2013,semeniuta2016analysis}. As in machine learning, testing errors significantly higher than training errors (calibration residuals) indicate an overfit which is directly related to an uncertainty in model parameters.
While a testing error is a valuable measure for the accuracy of a calibration, it requires capturing a full additional test dataset. 
In practice, this overhead can rarely be afforded which is why we focus on metrics that rely only on the calibration data.

\section{Calibration framework}
\subsection{Camera projection modeling} \label{sec:camera_projection}
From a geometric perspective, cameras project points in the 3D world to a 2D image~\cite{hartley_multiple_2004}. This projection can be expressed by a function $\boldsymbol{p}:\mathbb{R}^3\rightarrow\mathbb{R}^2$ that maps a 3D point $\boldsymbol{x}=(x, y, z)^T$ from a world coordinate system to a point $\bar{\boldsymbol{u}}=(\bar{u}, \bar{v})^T$ in the image coordinate system.
The projection can be decomposed into a coordinate transformation from the world coordinate system to the camera coordinate system $\boldsymbol{x} \rightarrow \boldsymbol{x_{c}}$ and the projection from the camera coordinate system to the image $\boldsymbol{p_C}:\boldsymbol{x_{c}}\rightarrow \bar{\boldsymbol{u}}$:
\begin{equation}
\bar{\boldsymbol{u}} = \boldsymbol{p}(\boldsymbol{x}, \boldsymbol{\theta}, \boldsymbol{\Pi}) = \boldsymbol{p_C}(\boldsymbol{x_{c}}, \boldsymbol{\theta}) = \boldsymbol{p_C}(\boldsymbol{R} \boldsymbol{x} + \boldsymbol{t}, \boldsymbol{\theta}), 
\end{equation}
where $\boldsymbol{\theta}$ are the \emph{intrinsic} camera parameters and $\boldsymbol{\Pi}$ are the \emph{extrinsic} parameters describing the rotation $\boldsymbol{R}$ and translation $\boldsymbol{t}$. For a plain pinhole model, the intrinsic parameters are the focal length $f$ and the principal point $(c_x, c_y)$, i.e. $\boldsymbol{\theta} = (f, c_x, c_y)$. For this case, the projection $\boldsymbol{p_C}(\boldsymbol{x_{c}}, \boldsymbol{\theta})$ is given by 
\begin{align}
\bar{u} =f/z_{c}\cdot x_{c} + c_x,\nonumber\\ 
\bar{v} = f/z_{c}\cdot y_{c} + c_y. 
\end{align}
To account for lens distortions, more complex models are needed.
In the following, we will consider a standard pinhole camera model with varying numbers of radial distortion parameters $k_1$, $k_2$, $k_3$ and different focal lengths $f_x$, $f_y$
\begin{align}
\bar{u} =f_x \frac{x_{c}}{z_{c}} (1 + k_1 r^2 + k_2  r^4 + k_3  r^6) + c_x,\nonumber\\ 
\bar{v} = f_y \frac{y_{c}}{z_{c}} (1 + k_1 r^2 + k_2  r^4 + k_3  r^6) + c_y,
\end{align}
where $ r = \sqrt{\left(\frac{x_{c}}{z_{c}}\right)^2+\left(\frac{y_{c}}{z_{c}}\right)^2}$. For more wide-angled lenses, we use the OpenCV fisheye model~\cite{opencv_fisheye}. 

\subsection{Calibration} \label{sec:calibration_framework}
Our methods build upon target-based camera calibration, in which planar targets are imaged in different poses realitive to the camera (see Fig.~\ref{fig:overview}A). Without loss of generality, we assume a single chessboard-style calibration target and a single camera in the following. The calibration dataset is a set of images $\mathcal{F}=\{\text{frame}_j\}_{j=1}^{N_\mathcal{F}}$. The chessboard calibration target contains a set of corners $\mathcal{C}=\{\text{corner}_i\}_{i=1}^{N_\mathcal{C}}$. The geometry of the target is well-defined, thus the 3D coordinates of chessboard-corner $i$ in the world coordinate system are known as $\boldsymbol{x}_{i}=(x_i, y_i, z_i)^T$. The image coordinates of chessboard-corners in each image are determined by a corner detection algorithm~\cite{strauss2014calibrating}, giving the observations $\boldsymbol{u}_i=(u_i, v_i)^T$. 
Depending on the corner detector, the perspective, the blur in the image, and many other factors, the observed coordinates $\boldsymbol{u}_i$ will deviate from the true image points $\boldsymbol{\bar{u}}_i$ by a certain error, called observational noise. This error is typically assumed to be independent identically distributed (i.i.d.) $\boldsymbol{\epsilon_d} \sim \mathcal{N}(\boldsymbol{0}, \sigma_d^2 \boldsymbol{I})$ with a certain detector variance $\sigma_d^2$. 
Previous work has used a more elaborate description of the detector error by taking into account the angle under which the corner is observed~\cite{peng_calibration_2019}, but we will stick to the most commonly used i.i.d. description.\par
Following~\cite{zhang_flexible_2000}, parameters are estimated by minimizing a calibration cost function, defined by the sum of squares of reprojection errors
\begin{equation}
\epsilon_{\mathrm{res}}^2 = \sum_{j\in \mathcal{F}} \sum_{i \in \mathcal{C}} ||\boldsymbol{u}_{ij} - \boldsymbol{p}(\boldsymbol{x}_{ij}, \boldsymbol{\theta}, \boldsymbol{\Pi}_j)||^2.
\end{equation}
For the sake of simplicity, we present formulas for non-robust optimization here. To reduce the impact of potential outliers, we advise robustification e.g. by using a Cauchy kernel. Optimization is peformed by a non-linear least-squares algorithm, which yields parameter estimates
$(\boldsymbol{\hat{\theta}}, \boldsymbol{\hat{\Pi}}) = \text{argmin}(\epsilon_{\mathrm{res}}^2)$.\\
To evaluate the calibration result, one of the most common metrics is the root mean squared error (RMSE) over all $N$ individual corners coordinates (observations) in the calibration dataset $\mathcal{F}$~\cite[p.133]{hartley_multiple_2004}:
\begin{equation}
\mathrm{RMSE}_{\mathrm{calib}} = \sqrt{\frac{1}{N} \sum_{j\in \mathcal{F}} \sum_{i \in \mathcal{C}}  ||\boldsymbol{u}_{ij} - \boldsymbol{p}(\boldsymbol{x}_{ij}, \boldsymbol{\hat{\theta}}, \boldsymbol{\hat{\Pi}}_j)||^2},
\end{equation}
or its squared version, the mean squared error $\mathrm{MSE}_{\mathrm{calib}}=\mathrm{RMSE}_{\mathrm{calib}}^2$. 
Note that the number of observations $N$ is twice the number of visible corners, i.e. $N=2 N_{\mathcal{C}} N_{\mathcal{F}}$ if the board was fully visible in all images.\par
The covariance matrix in estimated model parameters $\boldsymbol{\hat{\xi}}=(\boldsymbol{\hat{\theta}}, \boldsymbol{\hat{\Pi}})$ is theoretically given by the backpropagation of the covariance of the corner detector: 
\begin{equation}
\boldsymbol{\Sigma}_{\boldsymbol{\hat{\xi}}\boldsymbol{\hat{\xi}}} = ( \boldsymbol{J}_{\mathrm{calib}}^T  \boldsymbol{\Sigma}_{\boldsymbol{\epsilon_d \epsilon_d}}^{-1}  \boldsymbol{J}_{\mathrm{calib}})^{-1} = \sigma_d^2( \boldsymbol{J}_{\mathrm{calib}}^T  \boldsymbol{J}_{\mathrm{calib}})^{-1},
\end{equation}
where $\boldsymbol{\Sigma_{\boldsymbol{\epsilon_d \epsilon_d}}} = \sigma_d^2  \boldsymbol{I}$ is the covariance matrix of the corner detector and $\boldsymbol{J}_{\mathrm{calib}}$ is the Jacobian of calibration residuals~\cite[p.141-142]{hartley_multiple_2004}. 
As the variance of the corner detector $\sigma_d^2$ is not known a priori, it is typically replaced by the accuracy estimate $\hat{s}_d^2 = \mathrm{MSE}_{\mathrm{calib}}/(1-\frac{N_P}{N})$~\cite[p.92-96]{luhmann2013close}, giving
\begin{equation}
\hat{\boldsymbol{\Sigma}}_{\boldsymbol{\hat{\xi}}\boldsymbol{\hat{\xi}}, \mathrm{std}} = \hat{s}_d^2( \boldsymbol{J}_{\mathrm{calib}}^T  \boldsymbol{J}_{\mathrm{calib}})^{-1}.
\label{eq:cov}
\end{equation}
The covariance matrix contains the (co-)variances of both, the intrinsic parameters $\boldsymbol{\hat{\theta}}$ and extrinsic parameters $\boldsymbol{\hat{\Pi}}_j$ for each image $j$. The covariance matrix of intrinsic parameters $\boldsymbol{\Sigma_{\hat{\theta}\hat{\theta}}}$ can then be extracted as a submatrix of the full covariance matrix.

\begin{figure*}[ht!]
	\centering
	\includegraphics[width=0.99\textwidth]{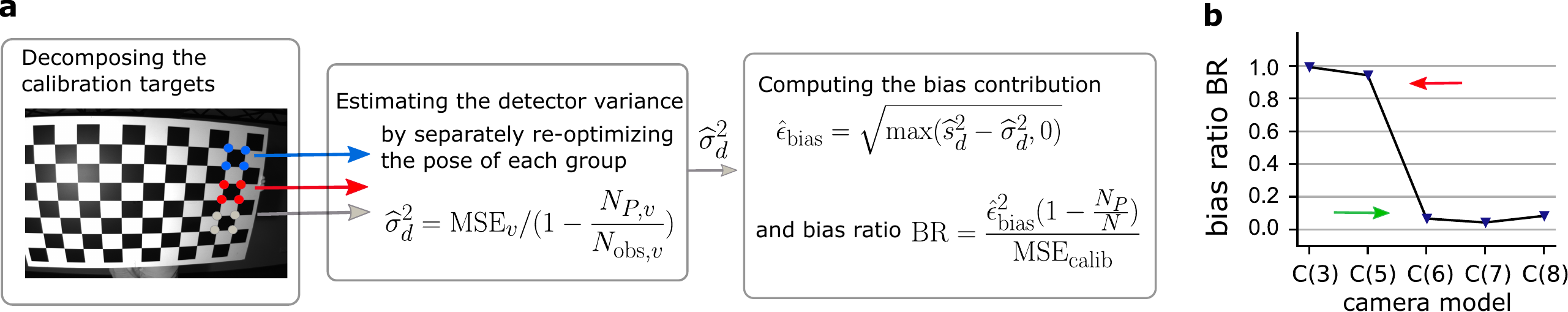}
	\caption{\textbf{Detecting systematic errors.} \textbf{a} Computation of the bias ratio: The calibration target is virtually decomposed into local groups, whose poses are re-optimized separately. Re-optimizating the poses can compensate a large fraction of potential systematic errors, so that the residuals provide an estimate for the random error contribution (detector variance $\sigma_d^2$). Using this estimate, the absolute bias $\epsilon_{\mathrm{bias}}$ and the BR can be computed. \textbf{b} Exemplary results of the BR for a simulated C(6) camera, when calibrating with models of increasing complexity. The BR clearly indicates the systematic error when using insufficiently complex models (C(3), C(5)).
	}
	\label{fig:biasMetric}
\end{figure*}
\section{Detecting Systematic Errors: The Bias Ratio}
The major challenge in quantifying systematic errors is the fact that the residuals 
are always a superposition of observational noise and potential systematic errors. As the magnitude of observational noise depends on multiple factors, including the camera, blur in the image, the calibration setup and many more, it is not known a priori. Consequently, the residuals do not provide direct information on the amount of bias. \\
In the following, we derive a novel method to disentangle these two contributions and
thereby quantify the fraction of systematic error in the mean squared reprojection error $\mathrm{MSE_{\mathrm{calib}}}$ of a calibration. 
 Following the assumptions made in Section \ref{sec:calibration_framework}, $\mathrm{MSE_{\mathrm{calib}}}$ can be formulated as a superposition of random errors, caused by the observational noise (detector variance $\sigma_d^2$), and a systematic error contribution.
Taking into account the number of parameters and the number of observations, one finds asymptotically (by augmentation of \cite[p. 136]{hartley_multiple_2004})
\begin{equation}
\mathrm{MSE_{\mathrm{calib}}} = \underbrace{\sigma_d^2 (1-\frac{N_\mathcal{P}}{N})}_{\substack{\mathrm{random \ error}}} + \underbrace{\ \epsilon_{\mathrm{bias}}^2 (1-\frac{N_\mathcal{P}}{N})\ }_{\substack{\mathrm{systematic\ error\ contribution}}},
\label{ML_RMSE}
\end{equation}
where $N_\mathcal{P}$ is the total number of free intrinsic and extrinsic parameters and $\epsilon_{\mathrm{bias}}$ denotes the bias introduced through systematic errors\footnote{Note, that the normalization of $\epsilon_{\mathrm{bias}}$ is different to \cite{hagemann2020bias}, to clarify its relation to the estimated accuracy $\hat s_d^2$. The definition of the bias ratio, however, remains unaffected.}. The detector variance $\sigma_d^2$ is generally camera-dependent and not known a priori. 
Thus, to disentangle random and systematic error contributions to $\mathrm{MSE_{\mathrm{calib}}}$, the detector variance $\sigma_d^2$ must be determined independently.\par
The rationale behind many calibration approaches, and in particular guided calibration, is to find most informative camera-target configurations. To estimate $\sigma_d^2$, we propose the opposite. We explicitly use configurations which are less informative for calibration but at the same time also less likely to be impacted by systematic errors. Such an uninformative configuration is given if a calibration target only covers small local image regions, as a large fraction systematic errors can then be compensated by adjusting the target pose. \par 
To obtain such an uninformative configuration without capturing additional data, we decompose the calibration target virtually into several smaller calibration targets $\mathcal{V} = \{\text{target}_v\}_{v=1}^{N_\mathcal{V}}$, usually consisting of exclusive sets of the four corners of a checker board tile (cf. Fig.~\ref{fig:biasMetric}\textbf{a}). The pose of each virtual calibration target in each image of the calibration dataset is then re-estimated individually while keeping the camera intrinsic parameters fixed. 
Here, pose estimation is overdetermined with a redundancy of two (four tile corners and six pose parameters). From the resulting mean squared errors, $\mathrm{MSE}_v$ with $v \in \mathcal{V}$, we compute estimates of $\sigma_d^2$ via~(\ref{ML_RMSE}). As systematic errors are mostly compensated after separately re-optimizing the poses, it can be assumed that $\epsilon^2_{\mathrm{bias}}$ is negligible here:
\begin{equation}\label{eq:sigmaests}
\widehat{\sigma}_{d_v}^2 = \frac{\mathrm{MSE}_v}{1-\frac{N_{P,v}}{N_{\mathrm{obs},v}}} = \frac{\mathrm{MSE}_v}{1-\frac{6}{8}} = 4~\mathrm{MSE}_v.
\end{equation}
To obtain an overall estimate of $\widehat{\sigma}_d^2$, we compute the MSE in (\ref{eq:sigmaests}) across the residuals of all virtual targets, using the median absolute deviation (MAD) as a robust estimator\footnote{Here, we assume the underlying distribution is Gaussian but might be subject to sporadic outliers. The MAD multiplied by a factor of $1.4826$ gives a robust estimate for the standard deviation \cite{rousseeuw}.}.\\
Given an estimate for the detector variance $\widehat{\sigma}_d^2$, we can use the decomposition of $\mathrm{MSE_{\mathrm{calib}}}$ (\ref{ML_RMSE}) to determine the systematic error contribution 
\begin{align}\label{eq:BAbs}
\hat{\epsilon}_{\mathrm{bias}}^2 &= \mathrm{max}\left( \frac{\mathrm{MSE}_{\mathrm{calib}}}{1-\frac{N_P}{N}}  -  \widehat{\sigma}_d^2, 0 \right) =\mathrm{max}(\widehat{s}_d^2 - \widehat{\sigma}_d^2, 0),
\end{align}
where $\mathrm{max}(\cdot,\cdot)$ ensures that the expression does not get negative due to the statistical nature of $\mathrm{MSE}_{\mathrm{calib}}$ and $\widehat{\sigma}_d^2$.
Finally, as a simple metric between zero and one, we compute the \textit{bias ratio} as
\begin{equation}
\mathrm{BR} =  \frac{\hat{\epsilon}_{\mathrm{bias}}^2(1-\frac{N_P}{N})}{\mathrm{MSE}_{\mathrm{calib}}}.
\end{equation}
The bias ratio is close to zero for unbiased calibration and close to one if the results are dominated by systematic errors. In combination, the bias ratio BR and the amount of bias $\hat{\epsilon}_{\mathrm{bias}}$ provide an intuitive measure for the systematic error in a calibration.\par
Note, that this method is closely related to an F-test for regression model comparison~\cite{doran1989applied}. By decomposing the calibration target, we are constructing a much more complex model. If there was no bias in the calibration, i.e. if the data was already explained without decomposing the target, the re-optimizations would only result in a further compensation of the observational noise. The reduction in the mean squared error (MSE) would then be fully explained by the larger number of free parameters. 
On the other hand, if the calibration was biased, the reduction in MSE would be larger than expected by merely adding unnecessary parameters. This additional reduction in MSE is the basis for computing the amount of bias $\epsilon_{\mathrm{bias}}$ and the bias ratio.
\par
Generally, this kind of analysis can be performed for any separable\footnote{The decomposition of the target must lead to an overdetermined estimation problem.} calibration target. In practice, one may also choose to cite $\sqrt{BR}$, as linear quantities may be easier to interpret than the quadratic BR.
The algrithm is summarized below.

\vspace{5mm}
\noindent
\fbox{\parbox{0.95\linewidth}{\textbf{Bias Ratio: Practical implementation}}}\\
\noindent
\fbox{\parbox{0.95\linewidth}{Computation of the bias ratio for target-based calibration procedures:
		\begin{enumerate}
			\item Perform robust camera calibration and extract a robust estimate of $\mathrm{MSE}_{\mathrm{calib}}$ and the optimal parameters $\boldsymbol{\hat{\theta}}$ and $\boldsymbol{\hat{\Pi}}$.
			\item Compute the residuals for all $v \in \mathcal{V}$:
			\begin{itemize}
				\item Decompose the calibration targets in each image into $N_\mathcal{V}$ exclusive virtual targets.
				\item Optimize their pose independently leaving $\boldsymbol{\hat{\theta}}$ unchanged.
			\end{itemize}
			\item Compute a robust estimate of the MSE over all residuals and determine $\widehat{\sigma}_d^2$ using (\ref{eq:sigmaests}).
			\item Use $\widehat{\sigma}_d^2$ to estimate the bias contribution $\epsilon_{\mathrm{bias}}$ via~(\ref{eq:BAbs}).
			\item Finally, compute the bias ratio as $\mathrm{BR} =  \hat{\epsilon}_{\mathrm{bias}}^2 (1-\frac{N_P}{N})/\mathrm{MSE}_{\mathrm{calib}}$.
\end{enumerate}}}

\section{Uncertainty Estimation}
\begin{figure*}[h!]
	\centering
	\includegraphics[width=0.85\textwidth]{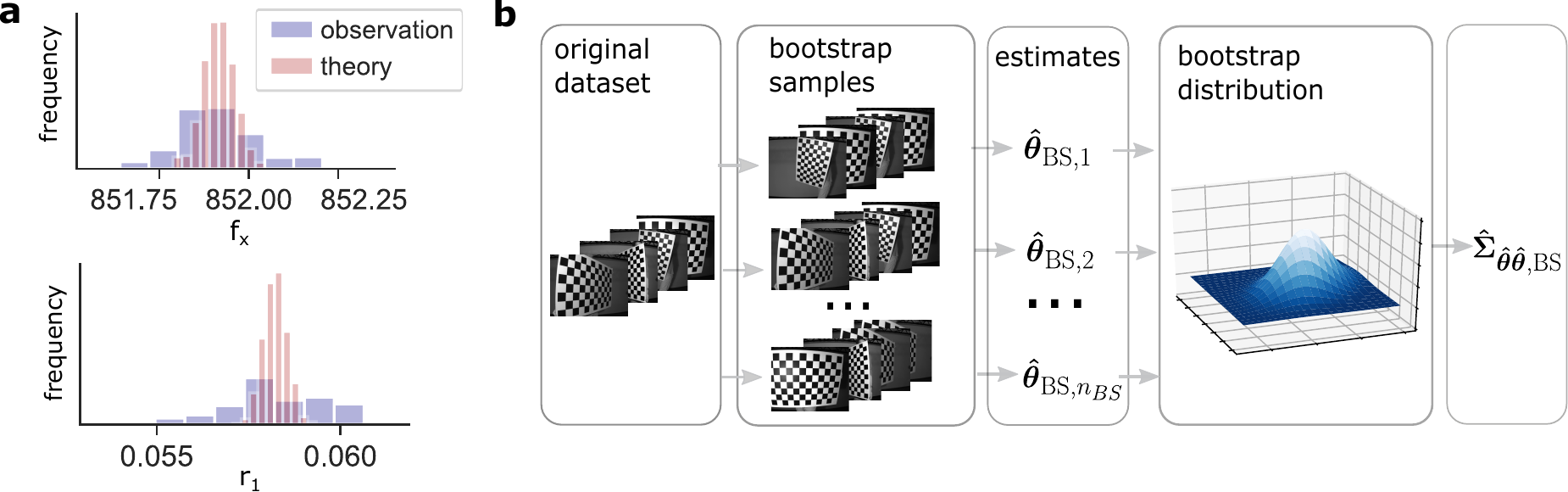}
	\caption{\textbf{Resampling-based uncertainty estimation.} \textbf{a} The standard estimator for the covariance matrix of model parameters underestimates the uncertainty in non-ideal, real-world scenarios. The histograms show the distribution of the focal length $f_x$ and radial distortion parameter $r_1$ when running calibrations with multiple image subsets of the same camera (manta lens). The observed variance is larger than predicted by the standard covariance estimator. \textbf{b} Resampling-based uncertainty estimation by bootstrapping the calibration dataset. Bootstrap samples are obtained by sampling images with replacement from the original dataset. For each bootstrap sample a calibration is performed, yielding a bootstrap distribution of estimates $\boldsymbol{\hat{\theta}}$. Finally, the covariance matrix is computed as the covariance of the bootstrap distribution. }
	\label{fig:BS_part1}
\end{figure*}
The second type of error, in addition to biases, are errors caused by a high uncertainty in estimated model parameters. The two major challenges in quantifying uncertainty are (i) the reliable estimation of the uncertainty and (ii) the formulation of parameter uncertainties as a model-independent and easily interpretable metric. In the following, we will address both challenges. First, we propose a resampling method to reliably estimate the covariance matrix of model parameters even under non-ideal conditions. Subsequently, we derive the model-independent uncertainty metric EME that predicts the expected error in image space.

\subsection{Resampling-based uncertainty estimation}
\begin{figure*}[ht!]
	\centering
	\includegraphics[width=0.99\textwidth]{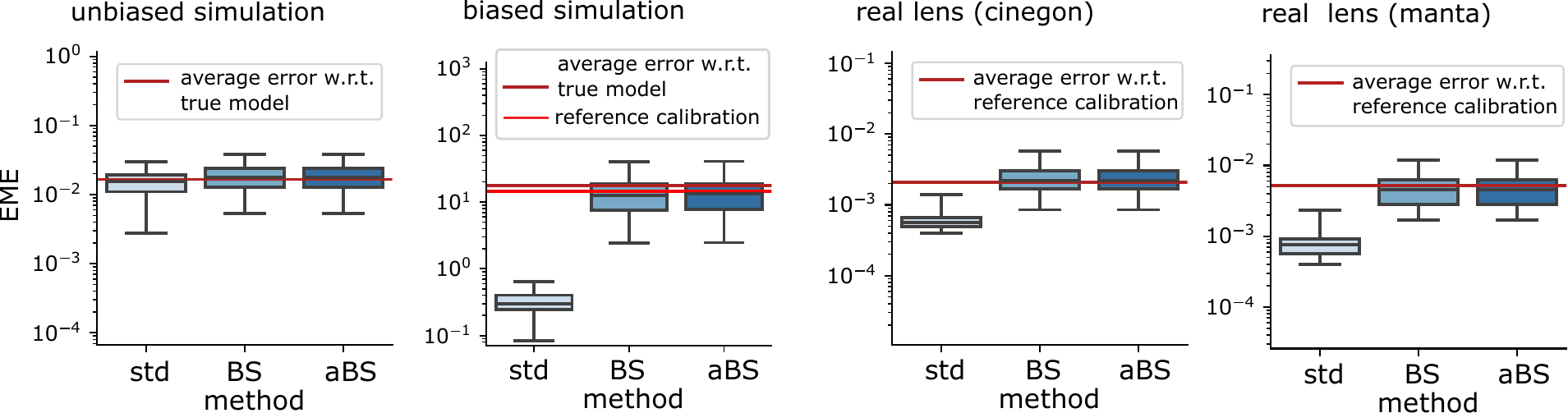}
	\caption{\textbf{Validation of resampling based uncertainty estimation.} For an ideal simulation, the standard method (std), the bootstrapping method (BS) and the approximated bootstrapping method (aBS) provide similar estimates for the uncertainty and the corresponding EMEs coincide with the average true error in image space (horizontal line). When simulating an underfit, the standard method significantly underestimated the uncertainty, while the estimates of both resampling methods remained close to the average error. For \emph{real cameras}, the results were similar to a simulated underfit. The standard method underestimated the uncertainty, while BS and aBS remained close to the true error. For comparability, same bootstrap samples were used for the BS and aBS method.}
	\label{fig:BS_part2}    
\end{figure*}

In general, the uncertainty of a calibration can be quantified by the covariance matrix of estimated model parameters. As parameters are estimated via nonlinear least squares, the standard estimator for the covariance matrix is a backpropagation of the detector variance (Eq.~\ref{eq:cov}). While this estimator is unbiased in theory and in simulations, we observed that for real data, it tends to \emph{underestimate} the uncertainty of a calibration. Comparing (a) the variance in estimated model parameters across multiple calibrations with (b) the average predicted variance, the latter tends to be smaller (see Figs.~\ref{fig:BS_part1}a,~\ref{fig:BS_part2}). 
This effect is especially detectable in the presence of small systematic errors, including imperfections of the calibration target or unmodeled lens distortion. As such small deviations from the ideal assumptions are almost inevitable in practice, a more robust estimator for the covariance matrix is needed to reliably estimate the uncertainty of a calibration. In the following, we will derive a resampling method to estimate uncertainty and we will propose an approximation of the method that is less computationally costly.

\subsection{Bootstrapping Method}
\label{sec:BS_method}
We propose to apply bootstrapping~\cite{davison1997bootstrap}, a nonparametric statistical technique, to obtain a more robust estimate of the covariance matrix (see Fig.~\ref{fig:BS_part1}b). Instead of relying on specific assumptions on the distribution of the data and the resulting analytical expressions, bootstrapping relies only on the data itself. The main assumption is that the dataset (sample) is representative for the population. Parameter variances and confidence intervals are then obtained by repeated resampling of the data. \par 
Let $\mathcal{F}=\{\text{frame}_j\}_{j=1}^{N_\mathcal{F}}$ be the set of $N_\mathcal{F}$ calibration images. 
We now construct $n_{\mathrm{BS}}$ bootstrap samples $\{\mathcal{F}_{\mathrm{BS},l}\}_{l=1}^{n_{\mathrm{BS}}}$ by sampling with replacement from the original set of images. Each of the bootstrap samples contains a total of $N_\mathcal{F}$ images (i.e. some images will be contained multiple times and others will be missing).
Now, the calibration is conducted with each of these bootstrap samples, yielding an estimate $\boldsymbol{\hat{\theta}}_{\mathrm{BS},l}$ for each sample $\mathcal{F}_{\mathrm{BS},l}$. The whole set of bootstrap estimates $\{\boldsymbol{\hat{\theta}}_{\mathrm{BS},l}\}_{l=1,...,n_{\mathrm{BS}}}$ forms the so-called bootstrap distribution. Based on this distribution, the covariance matrix can directly be estimated:
\begin{align}
\boldsymbol{\hat{\Sigma}}_{\boldsymbol{\hat{\theta}\hat{\theta}},\mathrm{BS}} = {\begin{aligned}{{\operatorname {Cov}} (\boldsymbol{\hat{\theta}}_{\mathrm{BS}}, \boldsymbol{\hat{\theta}}_{\mathrm{BS}})}&={\begin{pmatrix}\sigma_{1}^{2}&\sigma_{12}&\cdots &\sigma_{1 N_{\boldsymbol{\theta}}}\\\\\sigma_{21}&\sigma_{2}^{2}&\cdots &\sigma_{2N_{\boldsymbol{\theta}}}\\\\\vdots &\vdots &\ddots &\vdots \\\\\sigma_{N_{\boldsymbol{\theta}} 1}&\sigma_{N_{\boldsymbol{\theta}} 2}&\cdots &\sigma_{N_{\boldsymbol{\theta}}}^{2}
\end{pmatrix}}\end{aligned}}.
\end{align}
with
\begin{align}
{\hat {\sigma }}_{m}^{2}&:={\frac {1}{n_{\mathrm{BS}}-1}}\sum \limits _{l=1}^{n_{\mathrm{BS}}}\left(\theta_{lm}-{\overline {\theta}}_{m}\right)^{2}\\
{\hat {\sigma }}_{nm}&:={\frac {1}{n_{\mathrm{BS}}-1}}\sum _{l=1}^{n_{\mathrm{BS}}}(\theta_{ln}-{\overline {\theta}}_{n})(\theta_{lm}-{\overline {\theta}}_{m}).
\end{align}

\subsection{Approximated Bootstrapping Method}
Our results indicate that the bootstrap method provides a reliable estimate of the covariance matrix (see Section~\ref{sec:results_BS}). However, a major drawback lies in the computation time, as the entire calibration has to be conducted $n_{\mathrm{BS}}$ times.\\
To reduce the computation time, we propose an approximated bootstrapping method (aBS) which re-uses the already computed Jacobian of calibration residuals. Instead of conducting the entire calibration with each of the bootstrap samples, the aBS method only conducts the \emph{last iteration} of the nonlinear optimization, as we will explain in detail below.\\
The bundle adjustment performed during camera calibration relies on non-linear least squares estimation. Optimal parameters are found iteratively by performing so-called Gauss-Newton steps:
\begin{align}
(\boldsymbol{J^T J})\boldsymbol{\Delta\theta} &= \boldsymbol{J^T r}\\
\boldsymbol{\theta}^{k+1} &= \boldsymbol{\theta}^{k} + \boldsymbol{\Delta\theta}
\end{align}
where $\boldsymbol{J}$ is the local Jacobian at $\boldsymbol{\theta}^{k}$ and $\boldsymbol{r}$ are the residuals at the current iteration $k$. 
Note, that for calibration, an augmented system is typically used (e.g. Levenberg-Marquardt), but for the aBS method we omit the augmentation and use the plain Gauss-Newton formulation.
At the end of the calibration, the optimal parameters $\boldsymbol{\hat{\theta}}$, the local Jacobian $\boldsymbol{J}$ and the residuals $\boldsymbol{r}$ are known, where the $k$-th row of the Jacobian corresponds to the local derivatives $k$-th residual.\\
This is where the approximated bootstrapping method is applied: As for the original bootstrap method, we construct $n_{\mathrm{BS}}$ bootstrap samples $\{\mathcal{F}_{\mathrm{BS},l}\}_{l=1,...,n_{\mathrm{BS}}}$ by sampling with replacement from the original set of images. For each of these bootstrap samples $\mathcal{F}_{\mathrm{BS},l}$, the Jacobian and the residual vector are re-composed, such that they contain only the observations of $\mathcal{F}_{\mathrm{BS},l}$. Thus, for each bootstrap sample, we get a re-composed pair $(\boldsymbol{J}_{\mathrm{BS},l}, \boldsymbol{r}_{\mathrm{BS},l})$. For instance, if the bootstrap sample $\mathcal{F}_{\mathrm{BS},l}$ contained the first image of the original dataset twice, the pair 
would be given by
\begin{align}
\mathbf{J}_{BS,l} &= \begin{pmatrix} \mathbf{J}_1  \\ \mathbf{J}_1  \\ \mathbf{J}_3 \\ \vdots \\ \mathbf{J}_{N_\mathcal{F}} \end{pmatrix}, \quad
\mathbf{r}_{BS,l} = \begin{pmatrix} \mathbf{r}_1  \\ \mathbf{r}_1  \\ \mathbf{r}_3 \\ \vdots\\ \mathbf{r}_{N_\mathcal{F}} \end{pmatrix} \label{eq:7},
\end{align}
where the entry $\mathbf{J}_j$ describes the rows of the Jacobian that corrsepond to image $j$ and $\mathbf{r}_j$ is the residual vector corresponding to image $j$.\\
To obtain the bootstrap estimates $\boldsymbol{\hat{\theta}}_{\mathrm{BS},l}$, only the last Gauss-Newton step is conducted:
\begin{align}
(\boldsymbol{J}_{\mathrm{BS},l}^T \boldsymbol{J}_{\mathrm{BS},l})\boldsymbol{\Delta\theta}_{\mathrm{BS},l} &= \boldsymbol{J}_{\mathrm{BS},l}^T \boldsymbol{r}_{\mathrm{BS},l}\\
\boldsymbol{\hat{\theta}}_{\mathrm{aBS},l} &= \boldsymbol{\hat{\theta}} + \boldsymbol{\Delta\theta}_{\mathrm{BS},l}.
\end{align}
Having performed this last step for each bootstrap sample, we again get a distribution of bootstrap estimates $\{\boldsymbol{\hat{\theta}}_{\mathrm{aBS},l}\}_{l=1,...,n_{\mathrm{BS}}}$ and the covariance matrix $\boldsymbol{\hat{\Sigma}}_{\boldsymbol{\hat{\theta}\hat{\theta}},\mathrm{aBS}}$ can be estimated as in \ref{sec:BS_method}.\\
Thus, in short, the aBS simplifies the BS method by re-using the Jacobian and residual vector of the original calibration, which are typically costly to compute.

\subsection{The Expected Mapping Error}

\begin{figure*}[t]
	\centering
	\includegraphics[width=0.99\textwidth]{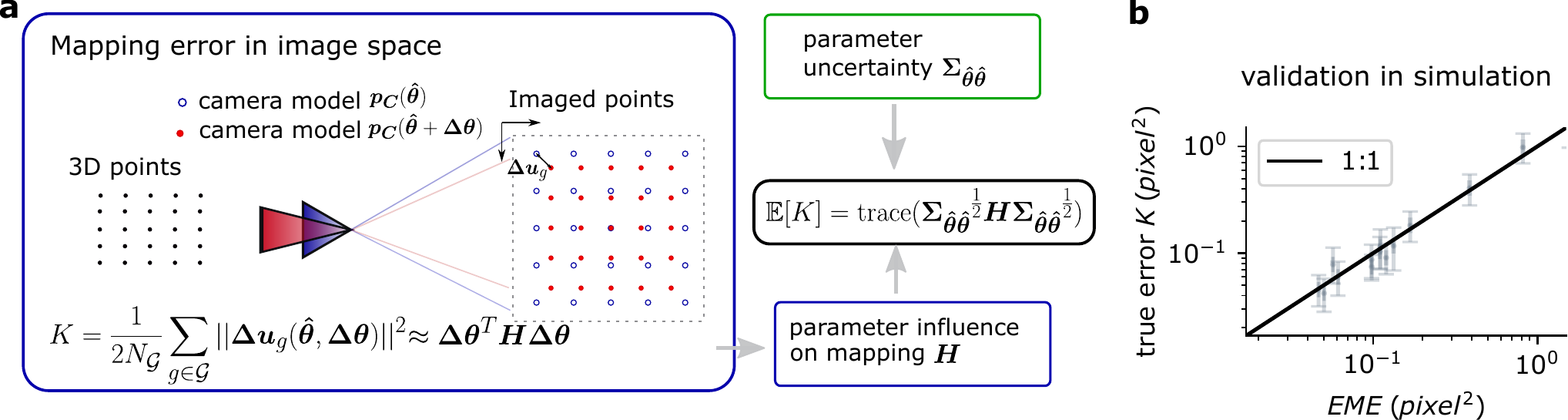}
	\caption{\textbf{Predicting the mapping error based on parameter uncertainties.} \textbf{a} Schematic of the derived uncertainty metric $\mathrm{EME}=\mathrm{trace}( \boldsymbol{\Sigma_{\hat{\theta}\hat{\theta}}}^{1/2} \boldsymbol{H}  \boldsymbol{\Sigma_{\hat{\theta}\hat{\theta}}}^{1/2})$. We define the mapping error $K$ as the difference between two camera models in image space. Approximating $K$ with a quadratic form in $\boldsymbol{\Delta\theta}$, we can predict its expected value by propagation of uncertainties.
		\textbf{b} Validation of the EME in simulations. The EME predicts the average true mapping error, i.e. the difference between the calibration result and the true camera model. Error bars are 95\% bootstrap confidence intervals.}
	\label{fig:uncertaintymetric}
\end{figure*}
Given a reliable estimate of the parameter covariance matrix, the second challenge lies in the formulation of the uncertainty in a model-independent and easily interpretable manner. 
We will now derive the \emph{expected mapping error} (EME), a novel uncertainty metric which quantifies the expected difference between the mapping of a calibration result $\boldsymbol{p_C}(\boldsymbol{x}, \boldsymbol{\hat{\theta}})$ and the true (unknown) model  $\boldsymbol{p_C}(\boldsymbol{x}, \boldsymbol{\bar{\theta}})$.\\
Inspired by previous works~\cite{cramariuc2020learning,richardson_aprilcal:_2013}, we quantify the mapping difference in image space, as pixel differences are easily interpretable. We define a set of points in image space $\mathcal{G}=\{\boldsymbol{u}_g\}_{g=1}^{N_\mathcal{G}}$,
from which the corresponding sight rays are computed via the inverse projection $\boldsymbol{p_C}^{-1}(\boldsymbol{u}_g, \boldsymbol{\bar{\theta}})$ using one set of model parameters $\boldsymbol{\bar{\theta}}$. Then, points on the viewing rays are backprojected to the image using the other set of model parameters $\boldsymbol{\hat{\theta}}$~\cite{beck_generalized_2018}.
The mapping error is then defined as the average squared distance between original image coordinates ${\boldsymbol{u}_g}$ and back-projected image points $\boldsymbol{p_C}(\boldsymbol{x}_g, \boldsymbol{\hat{\theta}})$ (see Fig.~\ref{fig:uncertaintymetric}):
\begin{equation}
\tilde{K}(\boldsymbol{\hat{\theta}}, \boldsymbol{\bar{\theta}}) = \frac{1}{2N_\mathcal{G}} \sum_{g\in\mathcal{G}} ||\boldsymbol{u_g} - \boldsymbol{p_C}(\boldsymbol{p_C}^{-1}(\boldsymbol{u_g}, \boldsymbol{\bar{\theta}}), \boldsymbol{\hat{\theta}})||^2,
\label{eq:mappingerrorsimple}
\end{equation}
where $2N_\mathcal{G}$ is the total number of image coordinates. \\
Formulation (\ref{eq:mappingerrorsimple}) of the mapping error assumes that an error in the intrinsics fully propagates to the image. However, the overall projection of a camera generally also includes the coordinate transformation from the world coordinate system to the camera coordinate system $\boldsymbol{x} \rightarrow \boldsymbol{x_{c}}$. As the camera pose and therefore the extrinsics are oftentimes re-estimated in practical settings, we additionally take into account that
 deviations in intrinsic parameters can partially be compensated by a change in extrinsic parameters~\cite{strauss_kalibrierung_2015} and allow for a virtual compensating rotation $\boldsymbol{R}$ of the viewing rays.
Thus, we formulate the \emph{effective} mapping error as follows:
\begin{equation} \label{eq:mappingerror}
K(\boldsymbol{\hat{\theta}}, \boldsymbol{\bar{\theta}}) = 
\underset{\boldsymbol{R}}{\text{min}}~
\frac{1}{2N_\mathcal{G}} \sum_{g\in\mathcal{G}} ||\boldsymbol{u}_g - \boldsymbol{p_C}(\boldsymbol{R}~ \boldsymbol{p_C}^{-1}(\boldsymbol{u}_g, \boldsymbol{\bar{\theta}}), \boldsymbol{\hat{\theta}})||^2.
\end{equation}
\par
The uncertainty metric EME is defined by the expected value of the mapping error, which can be reduced to a simple mathematical expression, as we will show below:
\begin{equation}\label{eq:EK}
\mathrm{EME}=\mathbb{E}[K(\boldsymbol{\hat{\theta}}, \boldsymbol{\bar{\theta}})]=\mathrm{trace}(\boldsymbol{\Sigma_{\hat{\theta}\hat{\theta}}}^{\frac{1}{2}} \boldsymbol{H} \boldsymbol{\Sigma_{\hat{\theta}\hat{\theta}}}^{\frac{1}{2}}),
\end{equation}
where $\boldsymbol{\Sigma_{\hat{\theta}\hat{\theta}}}$ is the covariance matrix in model parameters and $\boldsymbol{H}$ is the so-called \emph{model matrix} obtained from an approximation of the effective mapping error.\\
In the following, we will derive expression (\ref{eq:EK}).
Note, that the derivation is independent of the particular choice of $K$, provided that we can approximate $K$ with a Taylor expansion around $\boldsymbol{\hat{\theta}}=\boldsymbol{\bar{\theta}}$ up to second order:
\begin{equation}\label{eq:Kapprox}
\begin{split}
K(\boldsymbol{\hat{\theta}}, \boldsymbol{\bar{\theta}}) 
&\approx K(\boldsymbol{\bar{\theta}}, \boldsymbol{\bar{\theta}}) + \mathrm{grad}(K) \boldsymbol{\Delta\theta} + \frac{1}{2} \boldsymbol{\Delta\theta}^T \boldsymbol{H}_K \boldsymbol{\Delta\theta} \\
&\approx \frac{1}{2N_\mathcal{G}}\boldsymbol{\Delta \theta}^T  (\boldsymbol{J_{\mathrm{res}}}^T \boldsymbol{J_{\mathrm{res}}}) \boldsymbol{\Delta \theta} \\
&\approx \boldsymbol{\Delta \theta}^T \boldsymbol{H} \boldsymbol{\Delta \theta},
\end{split}
\end{equation}
where $\boldsymbol{\Delta \theta}=\boldsymbol{\bar{\theta}}- \boldsymbol{\hat{\theta}}$ is the difference between true and estimated intrinsic parameters, $\boldsymbol{\mathrm{res}}_g(\boldsymbol{\hat{\theta}}, \boldsymbol{\bar{\theta}})$ are the mapping residuals, i.e.
 $$\boldsymbol{\mathrm{res}}_g(\boldsymbol{\hat{\theta}}, \boldsymbol{\bar{\theta}}) = \boldsymbol{u}_g - \boldsymbol{p_C}(\boldsymbol{R}~ \boldsymbol{p_C}^{-1}(\boldsymbol{u}_g, \boldsymbol{\bar{\theta}}), \boldsymbol{\hat{\theta}}),$$ and $ \boldsymbol{J_{\mathrm{res}}}=d\boldsymbol{\mathrm{res}}/d\boldsymbol{\Delta\theta}$ is the Jacobian of the residuals. Furthermore, we defined the \emph{model matrix} $ \boldsymbol{H} := \frac{1}{2N_\mathcal{G}}\boldsymbol{J_{\mathrm{res}}}^T \boldsymbol{J_{\mathrm{res}}}$. 
For a more detailed derivation of the second step in~\ref{eq:Kapprox}, see Supplementary. \\
Estimated model parameters $\boldsymbol{\hat{\theta}}$ obtained from a least squares optimization are a random vector, asymptotically following a multivariate Gaussian with mean $\boldsymbol{\mu_{\theta}} = \boldsymbol{\bar{\theta}}$ and covariance $\boldsymbol{\Sigma_{\hat{\theta}\hat{\theta}}}$ ~\cite[p. 8]{goos_bundle_2000}. Likewise, the parameter error $\boldsymbol{\Delta \theta}=\boldsymbol{\bar{\theta}}- \boldsymbol{\hat{\theta}}$ follows a multivariate Gaussian, with mean $\boldsymbol{\mu_{\Delta\theta}} = \boldsymbol{0}$ and covariance $ \boldsymbol{\Sigma_{\Delta\theta \Delta\theta}} = \boldsymbol{\Sigma_{\hat{\theta}\hat{\theta}}}$. 
We propagate the distribution of the parameter error $\boldsymbol{\Delta \theta}$ to find the distribution of the mapping error $K(\boldsymbol{\hat{\theta}}, \boldsymbol{\bar{\theta}})$.
In short, we find that the mapping error $K(\boldsymbol{\hat{\theta}}, \boldsymbol{\bar{\theta}})$ can be expressed as a linear combination of $\chi^2$ random variables:
\begin{equation}\label{eq:Kderived}
\begin{split}
K(\boldsymbol{\hat{\theta}}, \boldsymbol{\bar{\theta}}) &= \boldsymbol{\Delta \theta}^T \boldsymbol{H} \boldsymbol{\Delta \theta}  \\
&= \sum_{n=1}^{N_{\boldsymbol{\theta}}} \lambda_n Q_n, \quad \text{with} \quad Q_n\sim \chi^2(1).
\end{split}
\end{equation}
The coefficients $\lambda_n$ are the eigenvalues of the matrix product $\boldsymbol{\Sigma_{\hat{\theta}\hat{\theta}}}^{\frac{1}{2}} \boldsymbol{H} \boldsymbol{\Sigma_{\hat{\theta}\hat{\theta}}}^{\frac{1}{2}}$ and $N_{\boldsymbol{\theta}}$ is the number of eigenvalues which equals the number of parameters $\boldsymbol{\theta}$. The full derivation of relation~(\ref{eq:Kderived}) is shown in the Supplementary. Importantly, based on expression~(\ref{eq:Kderived}), we can derive the expected value of  $K(\boldsymbol{\hat{\theta}}, \boldsymbol{\bar{\theta}})$:
\begin{equation}
\begin{split}
\mathbb{E}[K(\boldsymbol{\hat{\theta}}, \boldsymbol{\bar{\theta}})] &= \mathbb{E}[\sum_{n=1}^{N_{\boldsymbol{\theta}}} \lambda_n Q_n]  = \sum_{n=1}^{N_{\boldsymbol{\theta}}} \lambda_n \mathbb{E}[Q_n] = \sum_{n=1}^{N_{\boldsymbol{\theta}}} \lambda_n \\
&= \mathrm{trace}(\boldsymbol{\Sigma_{\hat{\theta}\hat{\theta}}}^{\frac{1}{2}} \boldsymbol{H} \boldsymbol{\Sigma_{\hat{\theta}\hat{\theta}}}^{\frac{1}{2}}),
\end{split}
\end{equation}
where we used that the $\chi^2$-distribution with one degree of freedom $\chi^2(1)$ has expectation value $\mathbb{E}[\chi^2(1)]=1$. 
We therefore propose the \emph{expected mapping error} $\mathrm{EME}=\mathrm{trace}(\boldsymbol{\Sigma_{\hat{\theta}\hat{\theta}}}^{\frac{1}{2}} \boldsymbol{H} \boldsymbol{\Sigma_{\hat{\theta}\hat{\theta}}}^{\frac{1}{2}})$ as a model-independent measure for the uncertainty.

\vspace{5mm}
\noindent
\fbox{\parbox{0.95\linewidth}{\textbf{Expected Mapping Error: Practical implementation}}}\\
\noindent
\fbox{\parbox{0.95\linewidth}{The expected mapping error $\mathrm{EME}$ can be determined for any given bundle-adjustment calibration: 
		\begin{enumerate}
			\item Run the calibration and extract the optimal parameters $\boldsymbol{\hat{\theta}}$ and the Jacobian $\boldsymbol{J}_{\mathrm{calib}}$ of the calibration cost function.
			\item Compute the intrinsic parameter covariance matrix $\boldsymbol{\Sigma_{\hat{\theta}\hat{\theta}}}$ using the resampling method, or, given ideal conditions, via the classical relation (\ref{eq:cov}).
			\item Determine the model matrix $ \boldsymbol{H}$:
			\begin{itemize}
				\item Implement the mapping error (Eq.~(\ref{eq:mappingerror})) as a function of the parameter estimate $\boldsymbol{\hat{\theta}}$ and a parameter difference $\boldsymbol{\Delta\theta}$.
				\item Numerically compute the Jacobian $\boldsymbol{J_{\mathrm{res}}} = d\boldsymbol{\mathrm{res}}/d\boldsymbol{\Delta\theta}$ at the estimated parameters $\boldsymbol{\hat{\theta}}$ and compute $\boldsymbol{H} = \frac{1}{2N_\mathcal{G}}\boldsymbol{J_{\mathrm{res}}}^T \boldsymbol{J_{\mathrm{res}}}$.
			\end{itemize}
			\item Compute $\mathrm{EME}=\mathrm{trace}( \boldsymbol{\Sigma_{\hat{\theta}\hat{\theta}}}^{\frac{1}{2}} \boldsymbol{H}  \boldsymbol{\Sigma_{\hat{\theta}\hat{\theta}}}^{\frac{1}{2}})$.
\end{enumerate}}}

\vspace{5mm}
Note, that although the EME is measured in the same units as the mean squared error $\mathrm{MSE}_{\mathrm{calib}}$, they conceptually differ. 
The $\mathrm{MSE}_{\mathrm{calib}}$ describes the deviation of observations in the calibration dataset from the model predictions. It can be interpreted as a \emph{training error} and could be reduced by simply using less observations or more model parameters in the calibration. The EME, on the other hand, can rather be interpreted as a prediction for the \emph{testing error}\footnote{More precisely, it is a prediction of the testing error, assuming a perfect test dataset where observations homogeneously cover the image and are free of noise. In a real test dataset, a random error of feature detection within the test images would have to be added.}. It will increase when less observations or more unneccessary model parameters are used and thereby indicate the potential overfit.

\label{sec:experiment}

\section{Experimental Evaluation}

\subsection{Simulations}
We simulated 3D world coordinates of a single planar calibration target in different poses relative to the camera (random rotations $\varphi_x, \varphi_y, \varphi_z \in [-\frac{\pi}{4}, \frac{\pi}{4}]$, translations $t_z\in[0.5~\text{m}, 2.5~\text{m}]$, $t_x, t_y\in[-0.5~\text{m}, 0.5~\text{m}]$). We then computed the resulting image coordinates using different camera models. To simulate the detector noise, we added Gaussian noise with $\sigma_d = 0.05$~px to all image coordinates.
For example images, see Fig.~\ref{fig:datasets}.

\subsection{Rendered images}
By simulating marker detection, we neglect the characteristics of the real corner, 
but assume i.i.d. Gaussian noise in the detection. To get a better understanding of potential deviations from this assumption, and their effect on our metrics, we additionally used rendered images. During rendering, we used a ray casting technique and simulated the camera's point spread function using multiple weighted samples per pixel. For rendered images, both the camera model and the calibration target are known and do not introduce any bias. Thereby, the full calibration pipeline, including corner detection, can be evaluated. For example images, see Fig.~\ref{fig:datasets}.

\subsection{Evaluation with real images}
\label{sec:experiment}
\begin{figure*}[h!]
	\centering
	\includegraphics[width=0.99\textwidth]{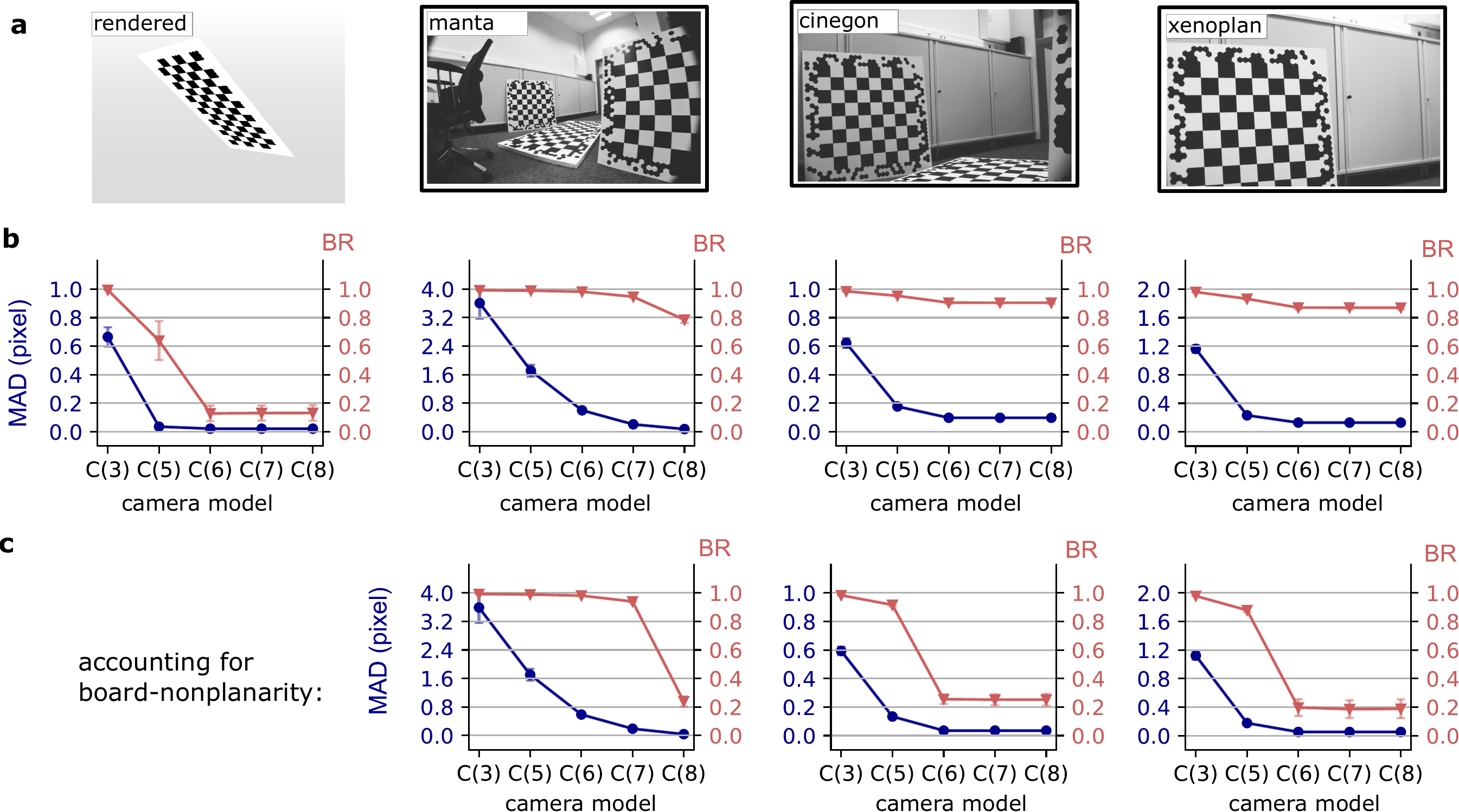}
	\caption{\textbf{Experimental evaluation of the bias ratio (BR).} \textbf{a} Exemplary rendered image and images of the same scene taken with the three different lenses used for evaluation. \textbf{b} Robust estimate of the RMSE (MAD, blue) and bias ratio (red) of calibrations with models of increasing complexities. For the rendered camera images, the BR drops once a model with the capability to describe the simulated mapping is employed. For all real lenses, the BR remains unexpectedly high even when using the fisheye model C(8). \textbf{c} Taking into account that the calibration target was not perfectly planar ($\Delta z \sim 10^{-4}$~m) further reduced the BR significantly. Error bars are standard deviations of ten calibrations with 50 random images each.}
	\label{fig:BR_experiments}       
\end{figure*}

We tested the methods using real images from three different lenses (see Fig.~\ref{fig:BR_experiments}a for example images and Fig.~\ref{fig:datasets} for more details on the lenses and the dataset). For each lens, we collected a dataset with images of a single planar calibration target. We use a coded target proposed by \cite{strauss2014calibrating}, to be able to associate corners across images. 
As no ground-truth model was available for the real lenses, we used a reference calibration\footnote{As reference, we used the average of ten calibrations with 50 random images each.} as an approximate ground-truth to verify the evaluation methods.

\section{Results}\label{sec:valbr}

\subsection{Evaluating the bias ratio}
\subsubsection*{Validating the BR in simulations and experiments}
\begin{figure*}
	\centering
	\includegraphics[width=0.99\textwidth]{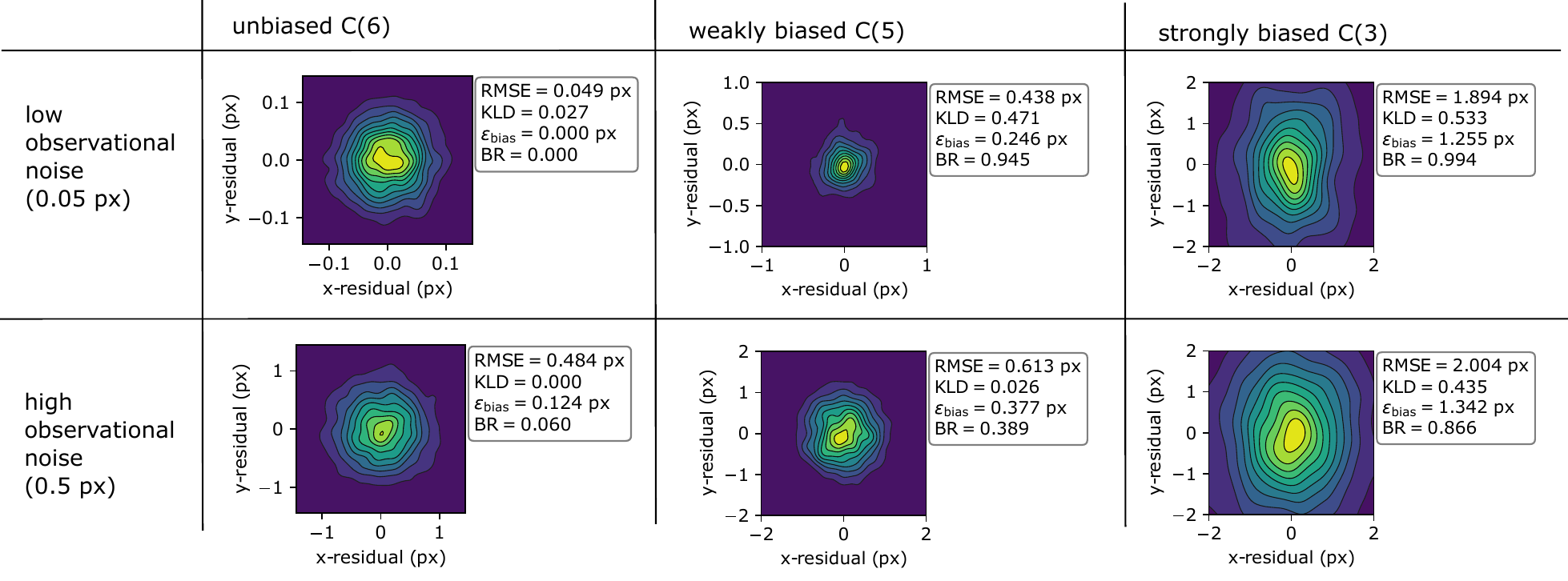}
	\caption{\textbf{Comparison of different bias metrics}. Each cell shows the distribution of residuals, the RMSE, the median Kullback-Leibler divergence (KLD), the bias ratio (BR) and the absolute bias ($\epsilon_{\mathrm{bias}}$) of the respective calibration. The results are based on simulated data, with varying degrees of observational noise and bias influences. Note the different limits of x- and y-axes. All metrics are of statistical nature, so that the precise values can vary for different calibrations. The same plots for real cameras are shown in the Supplementary.}
	\label{fig:BR_comparison}       
\end{figure*}
To test the bias metric, we ran calibrations using camera models of increasing complexity (increasing number of non-zero intrinsic parameters), including insufficiently complex models. We then computed the robust estimate of the RMSE (median absolute deviation, MAD) and the bias ratio (BR) for each calibration. For calibrations with an insufficiently complex model, the bias ratio should detect systematic errors.\par
Fig.~\ref{fig:BR_experiments} shows the results of calibrations of four cameras (one rendered dataset and three real cameras shown in Fig.~\ref{fig:BR_experiments}a). Each camera was calibrated with five different camera models of increasing complexity. 
In detail, the individual parameter sets are (cf. Sec. \ref{sec:camera_projection})
\begin{align}
\boldsymbol{\theta}_{C(3)} &= (f, c_x, c_y),\nonumber\\
\boldsymbol{\theta}_{C(5)} &= (f_x, f_y, c_x, c_y, k_1),\nonumber\\
\boldsymbol{\theta}_{C(6)} &= (f_x, f_y, c_x, c_y, k_1, k_2),\nonumber\\
\boldsymbol{\theta}_{C(7)} &= (f_x, f_y, c_x, c_y, k_1, k_2, k_3),\nonumber\\
\boldsymbol{\theta}_{C(8)} &= (f_x, f_y, c_x, c_y, k_1, k_2, k_3, k_4).\nonumber
\end{align}
For all cameras, the MAD and BR could be reduced by using more complex camera models, as the projections were not rectilinear and thus necessitate some kind of (nonlinear) distortion modeling.\par 
For the rendered images, the bias ratio was close to one for the first camera model C(3), indicating that the residuals were dominated by systematic errors. For the second camera model C(5), it reduced to $\mathrm{BR}\approx 0.6$, indicating that this model was better, but not yet sufficient. When using the camera model with two radial distortion parameters C(6), which is consistent with the model that was used for rendering, the BR dropped to $\mathrm{BR}< 0.2$. Adding additional parameters (C(7), C(8)) did not lead to a further reduction. The bias ratio thus correctly detected the presence of systematic errors for models C(3) and C(5) and indicated that one of the more complex models should be chosen for the calibration.\par
Surprisingly, the BR of all three real lenses remained comparatively high across all tested camera models, demonstrating that some sort of systematic error remained (Fig.~\ref{fig:BR_experiments}b). Further analyses revealed that the calibration board was not perfectly planar: running the calibrations again after precisely measuring the board geometry~\cite{strauss2014calibrating} lead to a significant further reduction in systematic error (Fig.~\ref{fig:BR_experiments}c). The non-planarity of the board was on the order of $10^{-4}$~m which demonstrates that even smallest imperfections in the calibration setup can be detected by the BR.\par
Using only the RMSE of the calibrations, some of the systematic errors could have been overlooked. For the rendered images, for instance, the RMSE already dropped to $\mathrm{RMSE}<0.1$~pixels when using camera model C(5), which may have been considered sufficiently low.
Likewise, even without taking into account the board-nonplanarity, the RMSE of all three real lenses was below 0.1 pixels and the calibrations may have been considered sufficiently accurate. This highlights the advantage of using the bias ratio to assess a calibration result.\par
For the calibrations shown in Fig.~\ref{fig:BR_experiments}, the BR got close to zero, but did not reach a value of zero for any of the rendered and real cameras. This could be explained by the fact that the observational noise of the true corner detector was not perfectly i.i.d., zero-mean Gaussian distributed~\cite{edwards2020experimental}. As the images used here were of high quality and generally showed a low observational noise ($\sigma_d \sim 0.03$~px), such small deviations from the expected Gaussian can become visible in the BR. For the real lenses, it is furthermore conceivable that some small systematic error remained, despite already taking into account the deformation of the calibration target.

\subsubsection*{Comparison of different bias metrics}
We compare the bias ratio with other state-of-the-art bias metrics introduced in Section~\ref{sec:stateoftheart}. We focus on the RMSE, the visualization of residuals in a smoothed 2D histogram~\cite{beck_generalized_2018} and the Kullback-Leibler divergence between a 2D normal distribution and the empirical distribution of residual vectors~\cite{schops_why_2019}. While \cite{schops_why_2019} uses 50x50 grid across the image and computes the KL-divergence in each cell, we use a 4x4 grid, as we are working with chessboard targets and significantly less observations. We calibrated simulated cameras with high and low observational noise, using the correct camera model (C(6), \emph{unbiased}), using a model with only one radial distortion parameter (C(5), \emph{weakly biased}) and using a plain pinhole model (C(3), \emph{strongly biased}) (see Fig.~\ref{fig:BR_comparison}). \par
The RMSE increased with increasing amount of bias. However, as the observational noise it not known a priori, it is hard to tell whether a value of $\mathrm{RMSE}\approx0.4$~px reflects a correspondingly high observational noise $\sigma_d\approx 0.4$~px or whether the observational noise is lower, and the RMSE reflects a bias (Fig.~\ref{fig:BR_comparison}, upper middle cell).\par
The 2D histogram of residuals was already more informative: visible deviations from an expected 2D Gaussian indicated a bias (Fig.~\ref{fig:BR_comparison}, right column, Fig.~\ref{fig:BR_comparison_cameras}). However, as the simulations show, even in the presence of biases, such deviations are not always visible (Fig.~\ref{fig:BR_comparison}, middle column).\par
Instead of relying on visual inspection, the Kullback-Leibler divergence \emph{measures} deviations from the expected Gaussian distribution: Higher values indicate higher deviations and therefore systematic errors. The KLD clearly indicated the increasing bias in the presence of low observational noise (Fig.~\ref{fig:BR_comparison}, upper row). However, in the presence of high observational noise, this noise partially overshadowed systematic errors (Fig.~\ref{fig:BR_comparison}, lower middle cell). In this high noise / weak bias case, the KLD remained close to zero despite the bias (see also Fig.~\ref{fig:BR_comparison_cameras} for calibrations with real lenses).\par

A similar effect was visible for the bias ratio: in the presence of high observational noise, the BR was lower, as it describes the \emph{fraction} of systematic error. To get a full picture on the amount of systematic error, we therefore recommend to take into account both, the BR and the absolute amount of bias $\epsilon_{\mathrm{bias}}$. The absolute amount of bias $\epsilon_{\mathrm{bias}}$ increased consistent with the increasing amount of bias, regardless of the amount of observational noise. In combination, the BR and $\epsilon_{\mathrm{bias}}$ were able to capture the systematic errors in all scenarios (see also Fig.~\ref{fig:BR_comparison_cameras} for calibrations with real lenses).\par

Note, that all metrics presented here are of statistical nature, and the precise values will fluctuate depending on the dataset and the optimization, even if the same model is chosen. 
Note also, that the KLD metric and the 2D histogram were originally designed for calibrations with highly complex camera models and significantly more observations~\cite{schops_why_2019,beck_generalized_2018}. It is therefore conceivable that these metrics are more powerful in such settings.\par


\subsection{Evaluating the resampling-based uncertainty estimation}
\label{sec:results_BS}
To test the resampling-based uncertainty estimation, we compared the estimated (co-)variances to the average deviation of the calibration result from the ground-truth parameters. If variances were estimated correctly, the variance of a parameter should, on average, reflect the parameter's mean squared deviation from the true parameter.\par
We compared the covariance estimates of the standard method (std), the novel bootstrapping method (BS) and the approximated bootstrapping method (aBS) (Fig.~\ref{fig:BS_part2}). 
Using both, simulated and real images, we ran calibrations with 50 random datasets containing $N_{\mathcal{F}}=25$ images each. 
For better comparability, we visualize the EME, i.e. the propagation of the covariance matrices to image space.\par 
For an ideal simulation with no systematic errors and i.i.d. Gaussian distributed observational noise, all three methods provided similar estimates (Fig.~\ref{fig:BS_part2}, left). For all methods, the average EME was close to the true average mapping error\footnote{Mapping error compared to the simulated ground-truth camera model.}. \par
When simulating an underfit, i.e. when calibrating a camera with only one radial distortion parameter although the ground-truth camera contained two non-zero radial distortion parameters, the methods differed. The standard method significantly underestimated the uncertainty and thereby the EME. This means that the standard estimator lead to overly optimistic assumptions on the uncertainty. The BS method and the aBS method, on the other hand, remained close to the true average error.\par
Importantly, for the two \emph{real} cameras, the results were similar to the underfit: the standard method underestimated the variance, leading to overly optimistic assumptions on the uncertainty. The BS method and the aBS method, on the other hand, remained close to the true error\footnote{Mapping error compared to the reference calibration.}. Our results thus indicate that in practice, where small systematic errors are almost inevitable\footnote{The calibrations shown in Fig.~\ref{fig:BS_part2} already take into account the board non-planarity.}, the resampling-based methods provide more reliable estimates for the uncertainty than the standard method.

\subsection{Evaluating the uncertainty metric}

\subsubsection*{Validating the prediction of the EME}
To validate the uncertainty metric EME in simulations, we simulated datasets with different numbers of images ($N_{\mathcal{F}}\in\{10, 15, ..., 50\})$ and ran calibrations using a sufficiently complex camera model. Fig.~\ref{fig:uncertaintymetric}b shows the uncertainty metric $\mathrm{EME}=\mathrm{trace}(\boldsymbol{\Sigma_{\hat{\theta}\hat{\theta}}}^{\frac{1}{2}} \boldsymbol{H} \boldsymbol{\Sigma_{\hat{\theta}\hat{\theta}}}^{\frac{1}{2}})$ and the average true mapping error compared to the ground-truth model. Consistent with Eq.~(\ref{eq:EK}), the EME predicted the average mapping error, i.e. the EME and the average mapping error were approximately equal. \par
Fig.~\ref{fig:comparison}a,b show the same plots for two real cameras. Here, we used a reference calibration as ground-truth and ran calibrations with different numbers of images ($N_{\mathcal{F}}\in\{10, 15, ..., 50\})$. In Fig.~\ref{fig:comparison}a, we used the standard covariance estimator in the computation of the EME. The EME is highly correlated with the average error, but its absolute values are significantly lower. The EME must therefore be interpreted as an upper bound for the precision that can be achieved based on a given dataset. \par
In Fig.~\ref{fig:comparison}b, we used the aBS method in the computation of the EME. Using the aBS method, the EME was consistent with the average real mapping error across most calibrations. Only for calibrations with few images ($N_{\mathcal{F}}\leq 15)$, the EME tended to \emph{overestimate} the true error. 
This is caused by the fact that the theory of bootstrapping is based on asymptotic assumptions (infinite sample sizes). Although it typically works well for smaller sample sizes as well, predictions become less reliable. In the case of camera calibration, this results in overly pessimistic uncertainty estimates. Note, however, that for most applications an overly pessimistic error estimate is to be preferred over an overly optimistic estimate of the standard covariance estimator.\par In summary, when using the standard covariance estimator, the EME provides an \emph{upper bound} to precision that can be achieved based on a given dataset. When using the aBS method to estimate the covariance, the EME can be interpreted as an actual prediction of the mapping error given a sufficient number of calibration images.

\subsubsection*{Comparison of different uncertainty metrics}
\begin{figure}
	\includegraphics[width=0.48\textwidth]{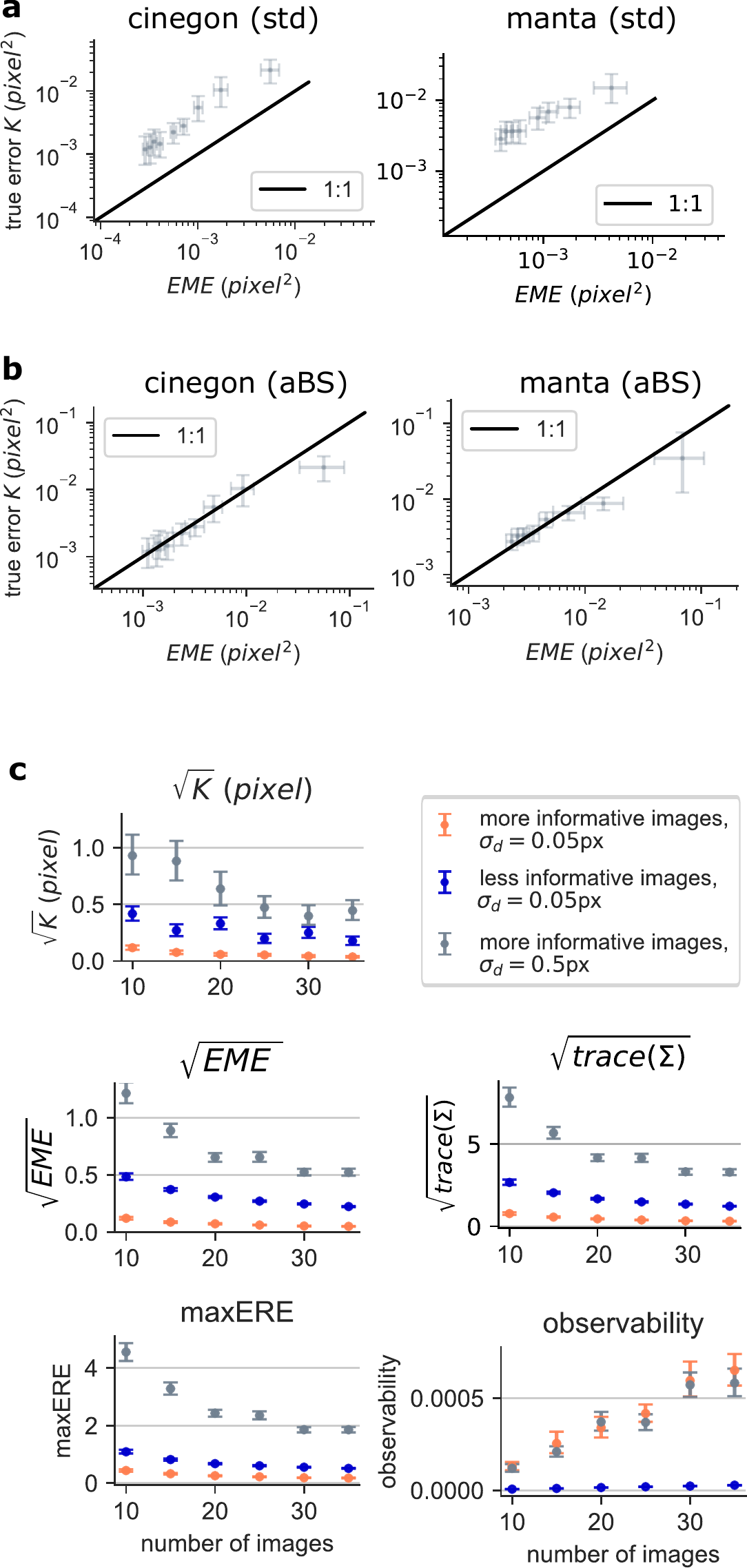}
	\caption{\textbf{Experimental evaluation of the expected mapping error (EME)}. \textbf{a, b} Validation of the EME with two real lenses (cinegon and manta). Using the standard covariance estimator (std), the EME and true error were highly correlated, but the absolute values of the EME were lower. Using the aBS method, the EME was consistent with the true average true error across most calibrations. \textbf{c} Comparison of state-of-the-art uncertainty metrics in simulated calibrations. On average, the true error $K$ decreases with the number frames. For comparability with \emph{maxERE}, we show $\sqrt{K}$ and $\sqrt{EME}$ in units of pixels. All metrics are correlated with the true error, but absolute values, scales, and units differ. Values are means across 50 random samples, error bars are 95\% bootstrap confidence intervals. }
	\label{fig:comparison}
\end{figure}
We compare the EME with the other state-of-the-art uncertainty metrics introduced in Section~\ref{sec:stateoftheart}. We focus on $\mathrm{trace}(\boldsymbol{\Sigma_{\hat{\theta}\hat{\theta}}})$, $maxERE$ \cite{richardson_aprilcal:_2013} and $observability$~\cite{strauss_kalibrierung_2015}, as these are the metrics closest to ours. Fig.~\ref{fig:comparison}\textbf{c} shows the result of all metrics for simulated calibrations with different numbers of images. To demonstrate how the uncertainty depends on both, the amount of observational noise $\sigma_d$ and the informativeness of calibration images, we simulated three different settings (low noise / informative images, low noise / less informative images, high noise / informative images).\par
Fig.~\ref{fig:comparison}\textbf{c} shows the true mapping error compared to the ground truth camera model $\sqrt{K}$, as well as all uncertainty metrics. The uncertainty, and therefore the average true error, decreased with an increasing number of images (an increasing number of observations $N$). As expected from theory, the scaling is approximately given by $\boldsymbol{\Sigma_{\hat{\theta}\hat{\theta}}} \propto N^{-1}$ (see also Fig.\ref{fig:comparison_sup}). Furthermore, the uncertainty was lowest for calibrations with low noise and informative images, and significantly higher for high noise or uninformative images.\par
All metrics were correlated with the true error. However, the metrics quantify uncertainty in very different ways: $\mathrm{trace}(\boldsymbol{\Sigma_{\hat{\theta}\hat{\theta}}})$ quantifies the uncertainty in model parameters, and thus inherently differs depending on the camera model. Furthermore, as its value represents the sum of different parameters variances, it does not have a unit. By re-parametrizing the camera model, the value of $\mathrm{trace}(\boldsymbol{\Sigma_{\hat{\theta}\hat{\theta}}})$ can change over orders of magnitude, making the comparison across different calibrations difficult.\par
The $observability$ metric shows a qualitatively different scaling than the other metrics as it increases linearly with the number of images. High observabilities imply that parameters were well observable based on the data. As the EME, the $observability$ accounts for the parameter's effect on the mapping and for compensations of errors in the intrinsics via different extrinsics. However, it does not incorporate the full uncertainty, but only the least observable direction. 
Furthermore, it does not account for the observational noise, but rather evaluates whether the setup was informative. It therefore serves to evaluate a calibration setup rather than the uncertainty of the specific calibration result.\par
Both, \emph{maxERE} and the EME quantify the expected error in image space and are thus easily interpretable. While \emph{maxERE} predicts a \emph{maximum} error, the EME reflects the \emph{average} error. In contrast to \emph{maxERE}, the EME does not require a Monte Carlo simulation. Furthermore, the EME can account for a compensation via different extrinsics, which we consider a reasonable assumption in most practical scenarios.

\section{Discussion and Conclusion}
In this work, we derived an evaluation scheme for camera calibration, including the detection of systematic errors and the quantification of uncertainty. \par
We have shown that it is possible to reliably capture systematic errors by disentangling the systematic errors from the random errors in the calibration residuals. The proposed method can thereby uncover wrong model assumptions and smallest imperfections in the calibration setup. 
Compared to other state-of-the-art methods to detect biases, the proposed method was more reliable and more easily interpretable, as it does not involve visual inspections of error distributions, nor does it require experience or comparative values from past calibrations.\par
To quantify uncertainty, we proposed a resampling-based estimation of the covariance matrix. In contrast to the standard covariance estimator, the resampling method does not impose strict assumptions on the error distribution and can therefore be applied under practical, non-ideal conditions. While the standard estimator led to an underestimation of the uncertainty in non-ideal settings, the proposed method provided reliable estimates even in the presence of systematic errors. One disadvantage of the resampling method is a required minimum size of the calibration dataset, otherwise the estimation becomes overly pessimistic. This is because the theory behind bootstrapping is based on asymptotic assumptions (infinite sample sizes), which are not met with too few calibration images. In most applications, however, such overly pessimistic error estimates are to be preferred over the overly optimistic estimates of the standard method.\par
To reduce the computation time of the resampling method, we additionally derived an approximation of the method that completely avoids recomputation of Jacobians and residuals and is therefore significantly faster. Thereby, the resampling method becomes feasible in practice. \par
Finally, we have derived a model-independent and easily interpretable uncertainty metric called \emph{expected mapping error} (EME). The EME quantifies the expected difference between the calibration result and the true camera in image space. By propagating the parameter uncertainty to image space, the metric is independent of the underlying camera model and comparable across different calibrations. The EME allows to quantify how informative a certain dataset is for the calibration and can thereby help to improve calibration setups.
\par
Importantly, our derivation of the EME can also be used to predict other, application specific errors. As a generic choice, we have used the average error across the image, but one could also weight image regions differently, or define an application-specific error, such as a triangulation error of a stereo system. Furthermore, the EME can be applied for calibration guidance, in order to guide users to collect calibration datasets that explicitly minimize the uncertainty~\cite{hagemann2020bias}.\par

In summary, our results suggest that target-based camera calibration can be evaluated reliably by
(i) computing the bias ratio to detect systematic errors, (ii) using the resampling method to estimate the uncertainty, and (iii) computing the EME to quantify the uncertainty in a model-independent manner. Together, the proposed methods can be used to assess the accuracy of individual calibrations, but also to benchmark new calibration algorithms, camera models, or calibration setups.

\begin{acknowledgements}
We thank Paul-Sebastian Lauer (Robert Bosch GmbH) for supporting the experimental setup and the data acquisition.
\end{acknowledgements}


\bibliographystyle{splncs04}
\bibliography{egbib}   
\newpage

\newcommand{\beginsupplement}{%
	\setcounter{table}{0}
	\renewcommand{\thetable}{S\arabic{table}}%
	\setcounter{figure}{0}
	\renewcommand{\thefigure}{S\arabic{figure}}%
	\setcounter{section}{0}
}

\onecolumn
\section*{Supplementary Material}
\beginsupplement

\begin{figure*}[h!]
	\centering
	\includegraphics[width=0.8\textwidth]{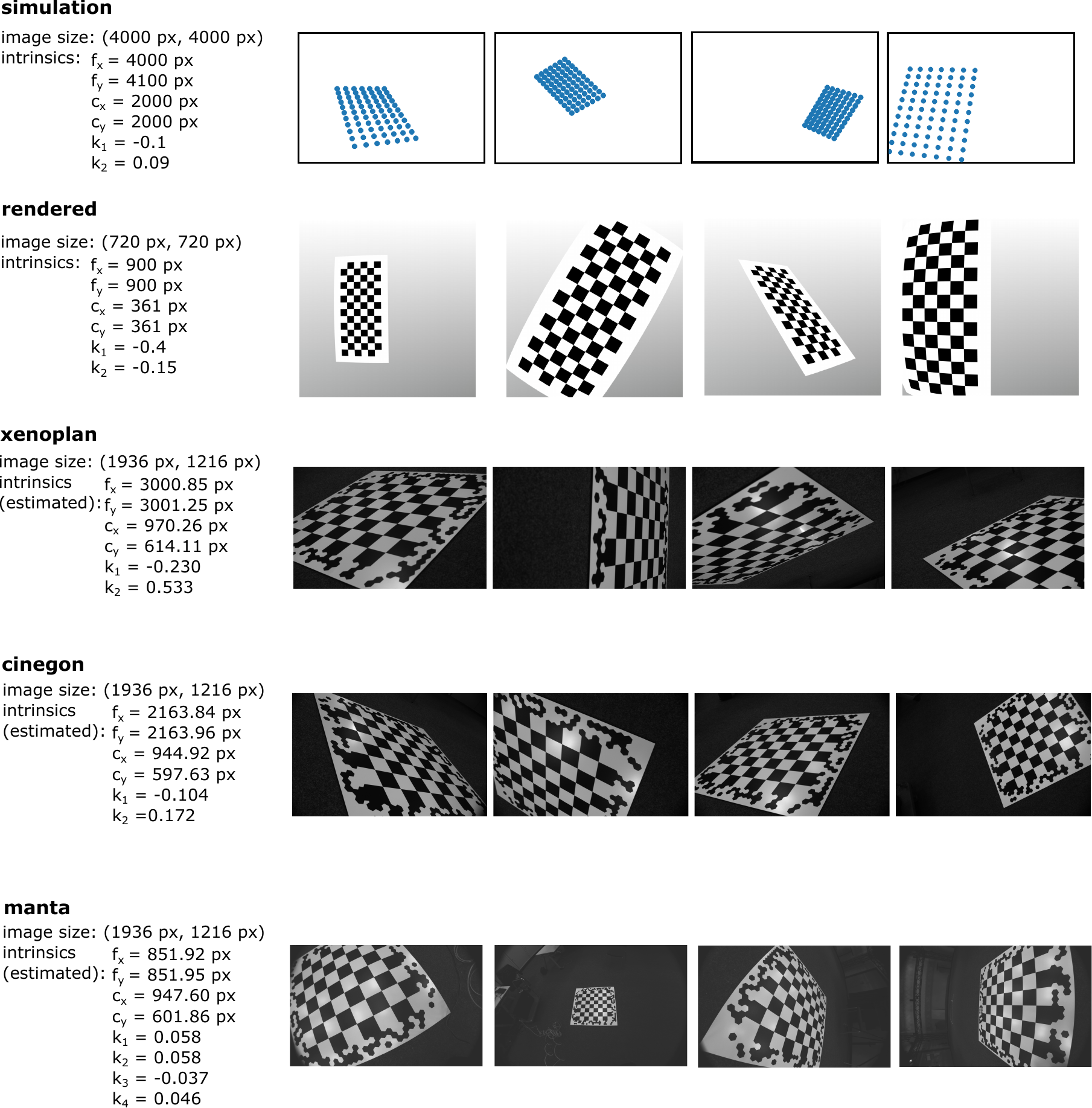}
	\vspace{4mm}
	\caption{\textbf{Example images of datasets used for evaluation.} For the real images (xenoplan, cinegon, manta), we used the same camera (Manta G-235 \cite{manta}), but different lenses (XENOPLAN 1.4/17, CINEGON 1.4/12, CINEGON 1.8/4.8).}
	\label{fig:datasets}       
\end{figure*}

\newpage
\FloatBarrier
\begin{figure*}
	\centering
	\includegraphics[width=0.95\textwidth]{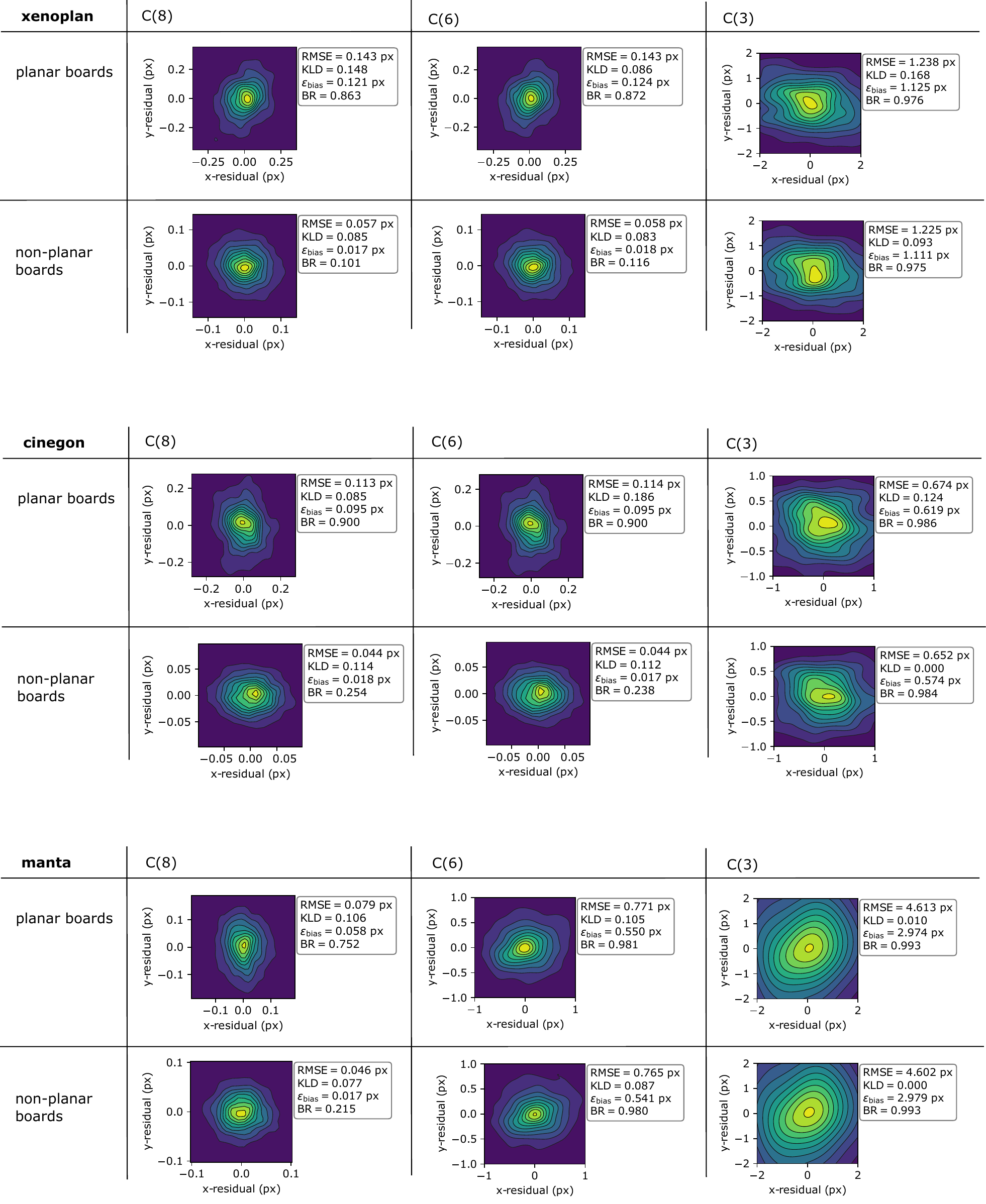}
	\caption{\textbf{Comparison of different bias metrics}. Each cell shows the distribution of residuals, the RMSE, the median Kullback-Leibler divergence (KLD), the bias ratio (BR) and the absolute bias ($\epsilon_{\mathrm{bias}}$) of a calibration.}
	\label{fig:BR_comparison_cameras}       
\end{figure*}
\FloatBarrier

\clearpage
\section{More detailed derivation of the expected mapping error (EME)}
In the following, we will show a more detailed derivation of the distribution of a quadratic mapping error $K(\boldsymbol{\hat{\theta}}, \boldsymbol{\bar{\theta}})$, which quantifies the difference between the mapping of a calibration result $\boldsymbol{p_C}(\boldsymbol{x}; \boldsymbol{\hat{\theta}})$ and the true (unknown) model  $\boldsymbol{p_C}(\boldsymbol{x}; \boldsymbol{\bar{\theta}})$.
The derivation is independent of the particular choice of $K(\boldsymbol{\hat{\theta}}, \boldsymbol{\bar{\theta}})$, provided that it can be approximated with a Taylor expansion around $\boldsymbol{\hat{\theta}}=\boldsymbol{\bar{\theta}}$ up to second order:
\begin{equation}
\begin{split}
K(\boldsymbol{\hat{\theta}}, \boldsymbol{\bar{\theta}}) &= \frac{1}{2N_\mathcal{G}}\boldsymbol{\mathrm{res}}(\boldsymbol{\hat{\theta}}, \boldsymbol{\bar{\theta}})^T \boldsymbol{\mathrm{res}}(\boldsymbol{\hat{\theta}}, \boldsymbol{\bar{\theta}}) \\
&\approx K(\boldsymbol{\bar{\theta}}, \boldsymbol{\bar{\theta}}) + \mathrm{grad}(K) \boldsymbol{\Delta\theta} + \frac{1}{2} \boldsymbol{\Delta\theta}^T \boldsymbol{H}_K \boldsymbol{\Delta\theta},
\end{split}
\end{equation}
where, $\boldsymbol{\mathrm{res}}(\boldsymbol{\hat{\theta}}, \boldsymbol{\bar{\theta}})$ are the mapping residuals and $2N_\mathcal{G}$ is the number of residuals. The gradient and the Hessian of $K$ w.r.t. the parameter error $\boldsymbol{\Delta\theta}=\boldsymbol{\hat{\theta}}-\boldsymbol{\bar{\theta}}$ are given by
\begin{equation}
\begin{split}
\mathrm{grad}(K) &= \frac{dK}{d\boldsymbol{\Delta\theta}}=2\frac{1}{2N_\mathcal{G}} \boldsymbol{\mathrm{res}}(\boldsymbol{\hat{\theta}}, \boldsymbol{\bar{\theta}})^T \boldsymbol{J_{\mathrm{res}}}\\
\boldsymbol{H}_K &= \frac{d^2K}{d\boldsymbol{\Delta\theta}^2} = 2 \frac{1}{2N_\mathcal{G}} \boldsymbol{J_{\mathrm{res}}}^T \boldsymbol{J_{\mathrm{res}}} + \text{higher order terms},
\end{split}
\end{equation}
where $ \boldsymbol{J_{\mathrm{res}}}=d\boldsymbol{\mathrm{res}}/d\boldsymbol{\Delta\theta}$ is the Jacobian of the residuals.
It follows that
\begin{equation}
\begin{split}
K(\boldsymbol{\hat{\theta}}, \boldsymbol{\bar{\theta}}) &\approx K(\boldsymbol{\bar{\theta}}, \boldsymbol{\bar{\theta}}) + \frac{2}{2N_\mathcal{G}} \boldsymbol{\mathrm{res}}(\boldsymbol{\bar{\theta}}, \boldsymbol{\bar{\theta}})^T \boldsymbol{J_{\mathrm{res}}} \boldsymbol{\Delta\theta}  + \frac{1}{4N_\mathcal{G}} \boldsymbol{\Delta\theta}^T (2\boldsymbol{J_{\mathrm{res}}}^T \boldsymbol{J_{\mathrm{res}}}) \boldsymbol{\Delta\theta}\\
&\approx \frac{1}{2N_\mathcal{G}}\boldsymbol{\Delta \theta}^T  (\boldsymbol{J_{\mathrm{res}}}^T \boldsymbol{J_{\mathrm{res}}}) \boldsymbol{\Delta \theta}\\
&\approx \boldsymbol{\Delta \theta}^T \boldsymbol{H} \boldsymbol{\Delta \theta},
\label{sup_eq:Kapprox}
\end{split}
\end{equation} 
where we defined the \emph{model matrix} $\boldsymbol{H} :=\frac{1}{2N_\mathcal{G}} \boldsymbol{J_{\mathrm{res}}}^T \boldsymbol{J_{\mathrm{res}}}$. Here we made the Gauss-Newton approximation $ d^2 \boldsymbol{\mathrm{res}}/d\boldsymbol{\Delta\theta}^2 \approx 0$, so that the higher-order terms of the Hessian vanish~\cite[p. 16]{goos_bundle_2000}. This approximation also implies that both $\boldsymbol{J_{\mathrm{res}}}$ and $\boldsymbol{H}$ are independent of where they are evaluated~\cite[p.16]{goos_bundle_2000}. 
In the second step we used that $\boldsymbol{\mathrm{res}}(\boldsymbol{\bar{\theta}}, \boldsymbol{\bar{\theta}})=\boldsymbol{0}$ and thus $K(\boldsymbol{\bar{\theta}},\boldsymbol{\bar{\theta}})=0$. This holds if we assume that the camera model is adequately chosen (especially not underparameterized).\\

\begin{figure}[h!]
	\centering
	\includegraphics[width=0.5\textwidth]{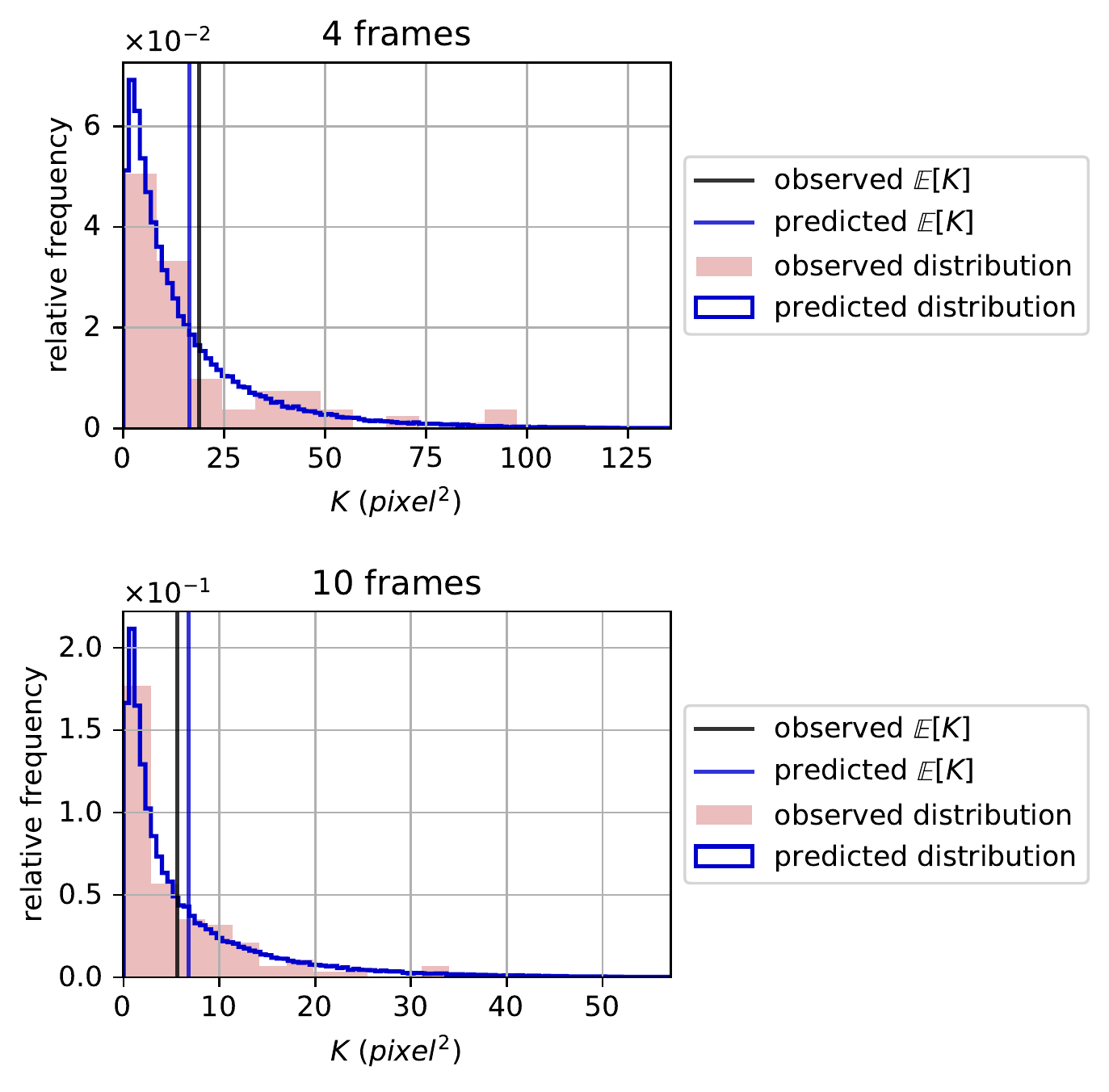}
	\caption{Predicted and observed distribution of the mapping error $K$ for simulated calibrations with $\sigma_d=1$~px. Vertical lines show true and predicted expected value $\mathbb{E}[K]$. With an increasing number of frames, the distribution becomes increasingly narrow (note the different x-axes). The prediction coincides well with the observed distribution.}
	\label{distributions}
\end{figure}

We now show that for an ideal, bias-free calibration, the effective mapping error $K(\boldsymbol{\hat{\theta}}, \boldsymbol{\bar{\theta}})$ can be predicted by propagating parameter uncertainties. Estimated model parameters $\boldsymbol{\hat{\theta}}$ obtained from a least squares optimization are a random vector, asymptotically following a multivariate Gaussian with mean $\boldsymbol{\mu_{\theta}} = \boldsymbol{\bar{\theta}}$ and covariance $\boldsymbol{\Sigma_{\hat{\theta}\hat{\theta}}}$. 
Likewise, the parameter error $\boldsymbol{\Delta \theta}$ follows a multivariate Gaussian, with mean $\boldsymbol{\mu_{\Delta\theta}} = \boldsymbol{0}$ and covariance $\boldsymbol{\Sigma_{\Delta\theta \Delta\theta}} = \boldsymbol{\Sigma_{\hat{\theta}\hat{\theta}}}$ (the quadratic cost function corresponds to the log-likelihood~\cite[p.9]{goos_bundle_2000}). 
Finally, as $\boldsymbol{\Delta \theta}$ is a random variable, so is $K(\boldsymbol{\hat{\theta}}, \boldsymbol{\bar{\theta}})$ a random variable.\\
We propagate the distribution of the parameter error $\boldsymbol{\Delta \theta}$ to find the distribution of the mapping error $K(\boldsymbol{\hat{\theta}}, \boldsymbol{\bar{\theta}})$: Based on $\boldsymbol{\Delta \theta}$, we can define a multivariate standard Gaussian distributed random variable
\begin{equation}
\begin{split}
\boldsymbol{z}&:=\boldsymbol{\Sigma_{\hat{\theta}\hat{\theta}}}^{-1/2}(\boldsymbol{\Delta \theta} - \boldsymbol{\mu_{\Delta\theta}}) \\
\Leftrightarrow \quad \boldsymbol{\Delta \theta} &= \boldsymbol{\Sigma_{\hat{\theta}\hat{\theta}}}^{1/2} \boldsymbol{z} + \boldsymbol{\mu_{\Delta\theta}} \\
\Leftrightarrow  \quad\boldsymbol{\Delta \theta}  &= \boldsymbol{\Sigma_{\hat{\theta}\hat{\theta}}}^{1/2} (\boldsymbol{z} + \boldsymbol{\Sigma_{\hat{\theta}\hat{\theta}}}^{-1/2}\boldsymbol{\mu_{\Delta\theta}}).
\end{split}
\end{equation}
Plugging this into the second order approximation of $K$ (\ref{sup_eq:Kapprox}), we get
\begin{equation}
\begin{split}
K &= \boldsymbol{\Delta \theta}^T \boldsymbol{H} \boldsymbol{\Delta \theta} \\
&=(\boldsymbol{z}+\boldsymbol{\Sigma_{\hat{\theta}\hat{\theta}}}^{-1/2}\boldsymbol{\mu_{\Delta\theta}})^T \boldsymbol{\Sigma_{\hat{\theta}\hat{\theta}}}^{\frac{1}{2}} \boldsymbol{H} \boldsymbol{\Sigma_{\hat{\theta}\hat{\theta}}}^{\frac{1}{2}} (\boldsymbol{z}+\boldsymbol{\Sigma_{\hat{\theta}\hat{\theta}}}^{-1/2}\boldsymbol{\mu_{\Delta\theta}}). 
\end{split}
\end{equation}
We now diagonalize the matrix $\boldsymbol{\Sigma_{\hat{\theta}\hat{\theta}}}^{\frac{1}{2}} \boldsymbol{H} \boldsymbol{\Sigma_{\hat{\theta}\hat{\theta}}}^{\frac{1}{2}} = \boldsymbol{P}^T \boldsymbol{\Lambda} \boldsymbol{P}$ where $\boldsymbol{\Lambda}$ is a diagonal matrix containing the eigenvalues and $\boldsymbol{P}$ is an orthogonal matrix ($\boldsymbol{P}^T \boldsymbol{P}=\boldsymbol{I}$) where the columns are the corresponding eigenvectors (spectral theorem).\\
Plugging this into the definition of $K$, we get
\begin{equation}
\begin{split}
K &=(\boldsymbol{z}+\boldsymbol{\Sigma_{\hat{\theta}\hat{\theta}}}^{-1/2}\boldsymbol{\mu_{\Delta \theta}})^T \boldsymbol{\Sigma_{\hat{\theta}\hat{\theta}}}^{\frac{1}{2}} \boldsymbol{H} \boldsymbol{\Sigma_{\hat{\theta}\hat{\theta}}}^{\frac{1}{2}} (\boldsymbol{z}+\boldsymbol{\Sigma_{\hat{\theta}\hat{\theta}}}^{-1/2}\boldsymbol{\mu_{\Delta\theta}}) \\
&= (\boldsymbol{z}+\boldsymbol{\Sigma_{\hat{\theta}\hat{\theta}}}^{-1/2}\boldsymbol{\mu_{\Delta\theta}})^T \boldsymbol{P}^T \boldsymbol{\Lambda} \boldsymbol{P} (\boldsymbol{z}+\boldsymbol{\Sigma_{\hat{\theta}\hat{\theta}}}^{-1/2}\boldsymbol{\mu_{\Delta\theta}})\\
&=  (\boldsymbol{P}\boldsymbol{z}+\boldsymbol{P}\boldsymbol{\Sigma_{\hat{\theta}\hat{\theta}}}^{-1/2}\boldsymbol{\mu_{\Delta\theta}})^T \boldsymbol{\Lambda}  (\boldsymbol{P}\boldsymbol{z}+\boldsymbol{P}\boldsymbol{\Sigma_{\hat{\theta}\hat{\theta}}}^{-1/2}\boldsymbol{\mu_{\Delta\theta}})\\
&= (\boldsymbol{u}+\boldsymbol{b})^T \boldsymbol{\Lambda} (\boldsymbol{u}+\boldsymbol{b}).
\end{split}
\end{equation}
In the last step, we replaced $\boldsymbol{b}=\boldsymbol{P}\boldsymbol{\Sigma_{\hat{\theta}\hat{\theta}}}^{1/2}\boldsymbol{\mu_{\Delta\theta}}$ and $\boldsymbol{u}=\boldsymbol{P}\boldsymbol{z}$. As $\boldsymbol{z}$ ist multivariate standard Gaussian distributed and $\boldsymbol{P}$ is orthogonal, $\boldsymbol{u}$ will also be multivariate standard Gaussian distributed with identity covariance and expectation zero.\\ 
Finally, since $\boldsymbol{\Lambda}$ is a diagonal matrix containing the eigenvalues $\lambda_{n=1,...,N_{\boldsymbol{\theta}}}$ of the matrix $\boldsymbol{\Sigma_{\hat{\theta}\hat{\theta}}}^{\frac{1}{2}} \boldsymbol{H} \boldsymbol{\Sigma_{\hat{\theta}\hat{\theta}}}^{\frac{1}{2}}$, this reduces to
\begin{equation}
\begin{split}
K &= \boldsymbol{\Delta \theta}^T \boldsymbol{H} \boldsymbol{\Delta \theta} \\
&= \sum_{i=n}^{N_{\boldsymbol{\theta}}} \lambda_n (u_n+b_n)^2 \quad \text{with} \quad u_n\sim \mathcal{N}(0, 1),
\label{K_derivedd}
\end{split}
\end{equation}
where $N_{\boldsymbol{\theta}}$ is the number of eigenvalues $\lambda_n$, which equals the number of parameters (the length of the parameter vector $\boldsymbol{\theta}$).\\
Expression~\ref{K_derivedd} is a linear combination of $\chi^2$-distributed random variables. The eigenvalues $\lambda_n$ are the coefficients and the $b_n$ determine the non-centrality.\\
As $\boldsymbol{\mu_{\Delta\theta}}=\boldsymbol{0}$, it follows that $\boldsymbol{b}=\boldsymbol{0}$. $K$ is therefore just a linear combination of central first-order $\chi^2$-distributed random variables:
\begin{equation}
K = \sum_{n=1}^{N_{\boldsymbol{\theta}}} \lambda_n Q_n, \quad \text{with} \quad Q_n\sim \chi^2(1).
\end{equation}
Fig.~\ref{distributions} shows examples of the distribution of the mapping error $K$, predicted by a single calibration result via equation~\ref{K_derivedd}, as well as the observed distribution, obtained from simulating $n_r=100$ noise realizations. The predicted and the observed distributions coincide well, which is consistent with our derivation.\\
There is no closed-form solution of the cumulative distribution function of the weighted sum of i.i.d. $\chi^2$-distributed random variables~\cite{bodenham2016comparison}.
However, it can be approximated, e.g. with the Imhof method~\cite{imhof1961computing}. Furthermore, the expected value is directly accessible:
The $\chi^2$-distribution with one degree of freedom $\chi^2(1)$ has expectation value $E[\chi^2(1)]=1$. We can thus determine the expected value of the mapping error $K$:
\begin{equation}
\begin{split}
\mathbb{E}[K(\boldsymbol{\hat{\theta}}, \boldsymbol{\bar{\theta}})] &= \mathbb{E}[\sum_{n=1}^{N_{\boldsymbol{\theta}}} \lambda_n Q_n] \nonumber = \sum_{n=1}^{N_{\boldsymbol{\theta}}} \lambda_n \mathbb{E}[Q_n] = \sum_{n=1}^{N_{\boldsymbol{\theta}}} \lambda_n \nonumber \\
&= \mathrm{trace}(\boldsymbol{\Sigma_{\hat{\theta}\hat{\theta}}}^{\frac{1}{2}} \boldsymbol{H} \boldsymbol{\Sigma_{\hat{\theta}\hat{\theta}}}^{\frac{1}{2}})\\
&= \mathrm{trace}(\boldsymbol{\Sigma_{\hat{\theta}\hat{\theta}}}\boldsymbol{H}).
\label{sup_eq:EK}
\end{split}
\end{equation}
This is the proposed uncertainty metric EME (expected mapping error).

\vspace{35mm}

\begin{figure}[h!]
	\centering
	\includegraphics[width=0.55\textwidth]{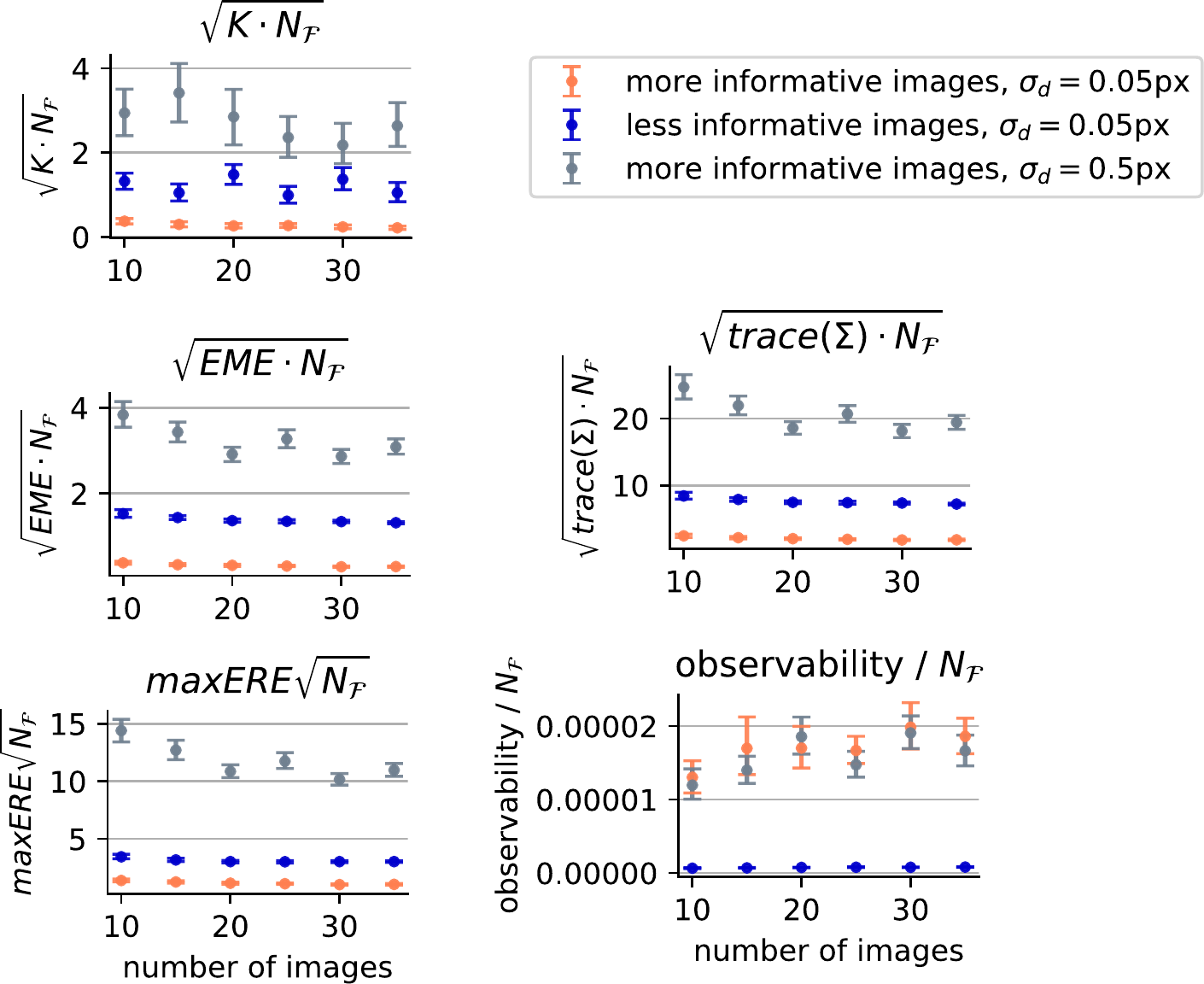}
	\caption{Re-scaled visualization of Fig.~\ref{fig:comparison}\textbf{c}. The uncertainty scales with the number of observations $N$, and therefore with the number of frames $N_{\mathcal{F}}$ approximately as $\boldsymbol{\Sigma_{\hat{\theta}\hat{\theta}}} \propto N^{-1}$. Re-scaling the metrics with the number of frames therefore gives approximately constant values. Values are means across 50 random samples, error bars are 95\% bootstrap confidence intervals. }
	\label{fig:comparison_sup}
\end{figure}
\clearpage

\section{Application in calibration guidance}
\begin{figure*}[!h]
	\centering
	\includegraphics[width=0.8\textwidth]{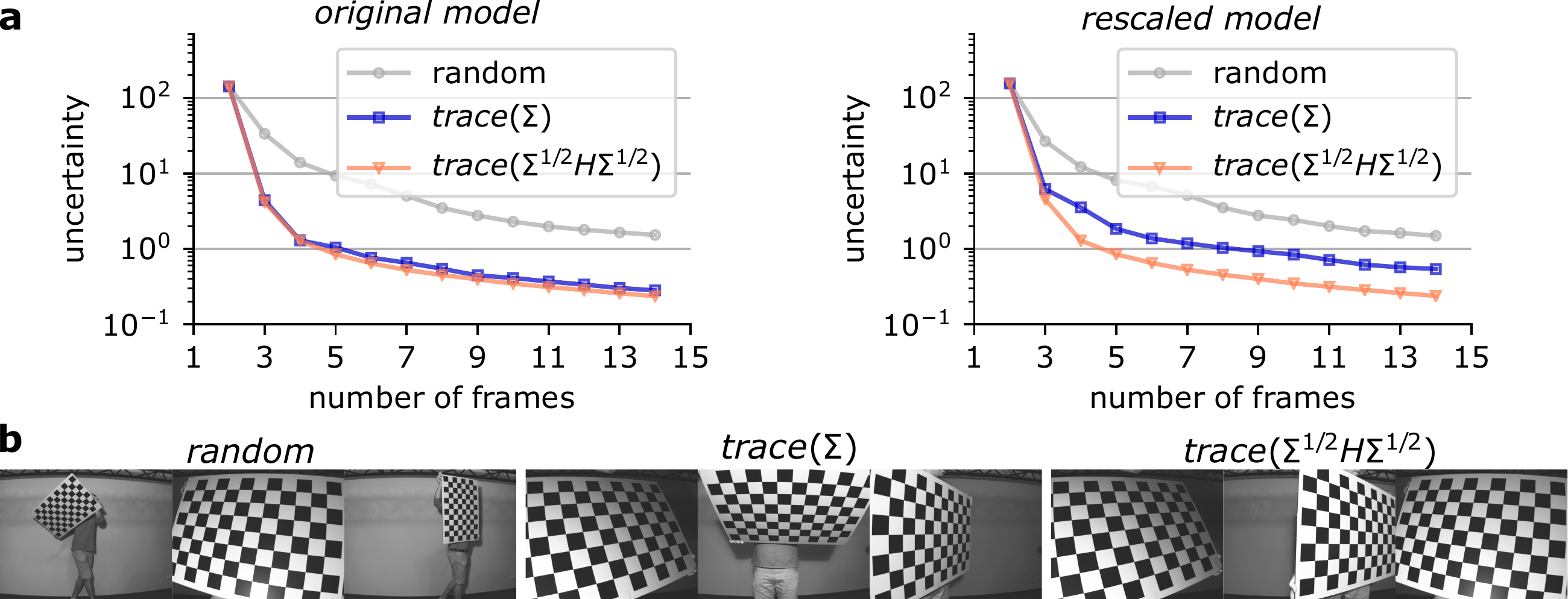}
	\caption{Application of EME for calibration guidance. 
		\textbf{a} For the original model, both metrics lead to a similarly fast reduction in uncertainty. Rescaling the model to a different unit of the focal length results in a reduced performance of $\mathrm{trace}(\boldsymbol{\Sigma_{\hat{\theta}\hat{\theta}}})$, while $\mathrm{trace}(\boldsymbol{\Sigma_{\hat{\theta}\hat{\theta}}}^{\frac{1}{2}} \boldsymbol{H} \boldsymbol{\Sigma_{\hat{\theta}\hat{\theta}}}^{\frac{1}{2}})$ remains unaffectd. Uncertainty is quantified by the average of the \emph{uncertainty map} proposed by calibration wizard~\cite{peng_calibration_2019}. \textbf{b} Examples of suggested poses.}
	\label{figure:wizard}
\end{figure*}
To demonstrate the practical use of the EME, we apply it in calibration guidance. Calibration guidance refers to systems that predict most informative next observations to reduce the remaining uncertainty and then guide users towards these measurements. We choose an existing framework, called calibration wizard~\cite{peng_calibration_2019} and extend it with our metric. \par
Calibration wizard predicts the next best pose by minimizing the trace of the intrinsic parameter's covariance matrix $\mathrm{trace}(\boldsymbol{\Sigma_{\hat{\theta}\hat{\theta}}})$. This is equivalent to minimizing the sum of the parameter's variances. However, depending on the camera model, parameters will affect the image in very different ways. High variance in a given parameter will not necessarily result in a proportionally high uncertainty in the image. \par
To avoid such an imbalance, we suggest to minimize the uncertainty in image space, instead of parameter space, i.e. to replace $\mathrm{trace}(\boldsymbol{\Sigma_{\hat{\theta}\hat{\theta}}})$ with the EME given by $\mathrm{trace}(\boldsymbol{\Sigma_{\hat{\theta}\hat{\theta}}}^{\frac{1}{2}} \boldsymbol{H} \boldsymbol{\Sigma_{\hat{\theta}\hat{\theta}}}^{\frac{1}{2}})$.\\
To compare the methods, we used a dataset of calibration images (see Fig.~\ref{figure:wizard}b). Starting with two random images, the system successively selected the most informative next image with (i) the original metric $\mathrm{trace}(\boldsymbol{\Sigma_{\hat{\theta}\hat{\theta}}})$, (ii) our metric $\mathrm{trace}(\boldsymbol{\Sigma_{\hat{\theta}\hat{\theta}}}^{\frac{1}{2}} \boldsymbol{H} \boldsymbol{\Sigma_{\hat{\theta}\hat{\theta}}}^{\frac{1}{2}})$ and (iii) randomly. 
Using the pinhole model with radial distortion, the poses suggested by $\mathrm{trace}(\boldsymbol{\Sigma_{\hat{\theta}\hat{\theta}}})$ and $\mathrm{trace}(\boldsymbol{\Sigma_{\hat{\theta}\hat{\theta}}}^{\frac{1}{2}} \boldsymbol{H} \boldsymbol{\Sigma_{\hat{\theta}\hat{\theta}}}^{\frac{1}{2}})$ were similarly well suited, both leading to a significantly faster reduction in uncertainty than random images (Fig.~\ref{figure:wizard}). 
However, when changing the camera model, e.g. by parameterizing the focal length in millimeters instead of pixels, simulated here by a division by 100 ($f \rightarrow 0.01\cdot f$), the methods differed:
the poses proposed by $\mathrm{trace}(\boldsymbol{\Sigma_{\hat{\theta}\hat{\theta}}}^{\frac{1}{2}} \boldsymbol{H} \boldsymbol{\Sigma_{\hat{\theta}\hat{\theta}}}^{\frac{1}{2}})$ reduced uncertainty significantly faster than $\mathrm{trace}(\boldsymbol{\Sigma_{\hat{\theta}\hat{\theta}}})$. This can be explained by the fact that when minimizing $\mathrm{trace}(\boldsymbol{\Sigma_{\hat{\theta}\hat{\theta}}})$, the variance of 
less significant parameters will be reduced just as much as the variance of parameters with large effect on the mapping. 
This example shows that the performance of $\mathrm{trace}(\boldsymbol{\Sigma_{\hat{\theta}\hat{\theta}}})$ can be affected by the choice of the model, while $\mathrm{trace}(\boldsymbol{\Sigma_{\hat{\theta}\hat{\theta}}}^{\frac{1}{2}} \boldsymbol{H} \boldsymbol{\Sigma_{\hat{\theta}\hat{\theta}}}^{\frac{1}{2}})$ remains unaffected.
\FloatBarrier

\end{document}